\documentclass[10pt,twocolumn,letterpaper]{article}

\usepackage[breaklinks=true,bookmarks=false]{hyperref}

\usepackage[utf8]{inputenc}
\usepackage{dirtytalk}
\usepackage{wacv}
\usepackage{times}
\usepackage{epsfig}
\usepackage{graphicx}
\usepackage{amsmath}
\usepackage{amssymb}
\usepackage{color}

\usepackage[caption=false,font=small,position=top,justification=centering]{subfig}
\usepackage{makecell}
\usepackage{multirow}
\usepackage[inline,shortlabels]{enumitem}
\usepackage{mathtools}
\usepackage{textcomp}
\usepackage{xcolor}
\usepackage{colortbl}
\usepackage{xspace}

\makeatletter
\@namedef{ver@everyshi.sty}{}
\makeatother
\usepackage{tikz}
\usepackage{pgfplots}

\usetikzlibrary{plotmarks,external}
\pgfplotsset{compat=newest}
\usepgfplotslibrary{patchplots}
\usepackage{caption,tabularx,booktabs}
\usepackage{cellspace}

\usepackage{float}

\newcolumntype{Y}{>{\centering\arraybackslash}X}
\usepackage[export]{adjustbox}

\newcommand{\secref}[1]{Sec.~{\ref{#1}}}
\newcommand{\figref}[1]{Fig.~{\ref{#1}}}
\newcommand{\tabref}[1]{Table~{\ref{#1}}}

\newcommand{\figsref}[1]{Figs.~{\ref{#1}}}

\newcommand{\mycaption}[2]{\caption[#1]{\emph{#1.} #2}}

\makeatletter
\DeclareRobustCommand\onedot{\futurelet\@let@token\@onedot}
\def\@onedot{\ifx\@let@token.\else.\null\fi\xspace}
 \def\Eg{\emph{E.g}\onedot}
\def\ie{\emph{i.e}\onedot}

\def\etal{\emph{et al}\onedot}
\makeatother

\newcommand{\RANSAC}{RANSAC\xspace}

\def\Gbs{Gr{\"o}bner bases\xspace}

\makeatletter
\newenvironment{customlegend}[1][]
{
    \begingroup
    \pgfplots@init@cleared@structures
    \pgfplotsset{#1}
}
{
  \pgfplots@createlegend
    \endgroup
}
\def\addlegendimage{\pgfplots@addlegendimage}
\makeatother

\newlength\fwidth
\DeclarePairedDelimiter{\diagfences}{(}{)}
\newcommand{\diag}{\operatorname{diag}\diagfences}

\newcommand{\inv}{{-1}}
\newcommand{\T}{{\!\top}}

\newcommand{\ma}[1]{\ensuremath{\mathtt{#1}}\xspace}

\newcommand{\mK}{\ensuremath{\ma{K}}\xspace}
\newcommand{\mR}{\ensuremath{\ma{R}}\xspace}

\newcommand{\ve}[1][x]{\ensuremath{\mathbf{#1}}\xspace}
\newcommand{\eve}[1][x]{\ensuremath{\boldsymbol{\mathsf{#1}}}\xspace}
\newcommand{\buildset}[3]{\ensuremath{ \{\,#1\,\}_{#2}^{#3} }\xspace}
\newcount\colveccount
\newcommand*\colvec[1]{
    \global\colveccount#1
    \begin{pmatrix}
    \colvecnext
}
\def\colvecnext#1{
    #1
    \global\advance\colveccount-1
    \ifnum\colveccount>0
        \\
        \expandafter\colvecnext
    \else
        \end{pmatrix}
    \fi
}
\newtoks\rowvectoks
\newcommand{\rowvec}[2]{
  \rowvectoks={#2,}\count255=#1\relax
  \advance\count255 by -1
  \rowvecnexta}
\newcommand{\rowvecnexta}{
  \ifnum\count255>0
    \expandafter\rowvecnextb
  \else
    \setlength\arraycolsep{1pt}     
    \begin{pmatrix}\the\rowvectoks\end{pmatrix}
  \fi}
\newcommand\rowvecnextb[1]{
  \ifnum\count255>1     
    \rowvectoks=\expandafter{\the\rowvectoks&#1,}
  \else
    \rowvectoks=\expandafter{\the\rowvectoks&#1}
  \fi
    \advance\count255 by -1
    \rowvecnexta}

\newcommand{\vX}[1][]{\ensuremath{\ve[X]_{#1}}\xspace}

\newcommand{\vx}[1][]{\ensuremath{\ve[x]_{#1}}\xspace}

\newcommand{\vxd}[1][]{\ensuremath{\ve[\tilde{x}]_{#1}}\xspace}
\newcommand{\vxdp}[1][]{\ensuremath{\ve[\tilde{x}]^{\prime}_{#1}}\xspace}

\newcommand{\xd}[1][]{\ensuremath{\tilde{x}_{#1}}\xspace}
\newcommand{\yd}[1][]{\ensuremath{\tilde{y}_{#1}}\xspace}

\newcommand{\rd}[1][]{\ensuremath{\tilde{r}_{#1}}\xspace}

\newcommand{\vm}[1][]{\ensuremath{\ve[m]_{#1}}\xspace}
\newcommand{\vmp}[1][]{\ensuremath{\vm[#1]^{\prime}}\xspace}
\newcommand{\vmd}[1][]{\ensuremath{\ve[\tilde{m}]_{#1}}\xspace}

\newcommand{\vu}[1][]{\ensuremath{\ve[u]_{#1}}\xspace}

\newcommand{\vl}[1][]{\ensuremath{\ve[l]_{#1}}\xspace}

\newcommand{\vU}[1][]{\ensuremath{\ve[U]_{#1}}\xspace}
\newcommand{\vV}[1][]{\ensuremath{\ve[V]_{#1}}\xspace}

\newcommand{\WildBMVC}{\textbf{5CA}\xspace}
\newcommand{\AntunCVPR}{\textbf{7CA}\xspace}
\newcommand{\EVP}{\textbf{4PC (EVP)}\xspace}
\newcommand{\EVL}{\textbf{6PC (EVL)}\xspace}

\newcommand{\HybridSolver}{\textbf{4PC+2CA}\xspace}
\newcommand{\HybridSolverTwo}{\textbf{2PC+4CA}\xspace}
\newcommand{\CircleSolver}{\textbf{6CA}\xspace}
\newcommand{\CircleDegSolver}{\textbf{5CA*}\xspace}

\newcommand{\CircHyb}{\CircleSolver \& \textbf{2PC+4CA}\xspace}
\newcommand{\AllProposed}{All}

\newcommand{\rmswarperr}{\ensuremath{\Delta^{\mathrm{RMS}}}\xspace}
\definecolor{mycolor1}{rgb}{0,0,0}
\definecolor{mycolor2}{rgb}{1,0,1}
\definecolor{mycolor3}{rgb}{0,1,1}
\definecolor{mycolor4}{rgb}{0,0,1}
\definecolor{mycolor5}{rgb}{0,1,0}
\definecolor{betteryellow}{rgb}{1,0.8824,0.0980}
\definecolor{lavender}{rgb}{0.902,0.7451,1.0}
\definecolor{olive}{rgb}{0.5020,0.5020,0} 
\definecolor{orange}{rgb}{1,0.5,0} 
\definecolor{bettergreen}{rgb}{0,0.6,0.3}

\definecolor{blue}{HTML}{4069B0}
\definecolor{lightorange}{HTML}{FF8F00}
\definecolor{orange}{HTML}{E45611}
\definecolor{darkorange}{HTML}{B85325}
\definecolor{lightgreen}{HTML}{34DC5B}
\definecolor{green}{HTML}{28A745}
\definecolor{lightgray}{HTML}{656972}
\definecolor{gray}{HTML}{53585F}
\definecolor{darkgray}{HTML}{333333}
\definecolor{red}{HTML}{D61901}
\definecolor{magenta}{HTML}{CC00CC}
\definecolor{cyan}{HTML}{00FFFF}

\colorlet{EVP}{bettergreen}
\colorlet{WildBMVC}{mycolor2}
\definecolor{HybridSolver}{rgb}{1,0,0}
\definecolor{HybridSolverTwo}{rgb}{0,0,0}
\definecolor{CircleDegSolver}{HTML}{4069B0}
\definecolor{CircleSolver}{rgb}{0,1,1}
\colorlet{CircHyb}{betteryellow}
\colorlet{AllProposed}{lavender}

\wacvfinalcopy

\begin{document}

\title{Minimal Solvers for Single-View Lens-Distorted Camera Auto-Calibration}
\author{Yaroslava Lochman$^{1,2}$\\
{\tt\small lochman@ucu.edu.ua}
\and
Oles Dobosevych$^{1}$\\
{\tt\small dobosevych@ucu.edu.ua}
\and
Rostyslav Hryniv$^{1}$\\
{\tt\small rhryniv@ucu.edu.ua}
\and
James Pritts$^{2}$\\
{\tt\small jbpritts@fb.com}
\and
$^{1}$Machine Learning Lab, Ukrainian Catholic University in Lviv
\and
$^{2}$Facebook Reality Labs in Pittsburgh\\
}

\makeatletter
\g@addto@macro\@maketitle{
\vspace{-35pt}
\begin{figure}[H]
\setlength{\linewidth}{\textwidth}
\setlength{\hsize}{\textwidth}
\newlength{\h}
\setlength{\h}{0.236\columnwidth}
\centering
\captionsetup[subfigure]{}
\subfloat[Input]{
\includegraphics[height=\h]{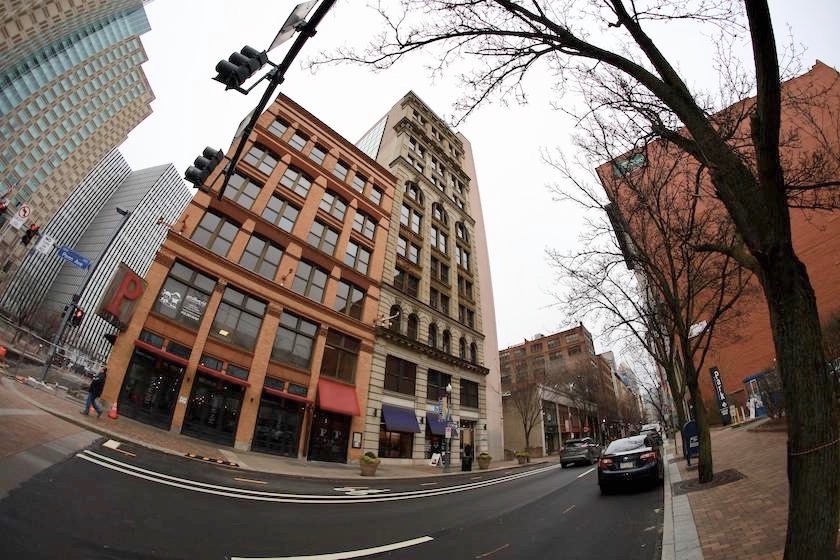}}
\subfloat[Undistorted]{
\includegraphics[height=\h]{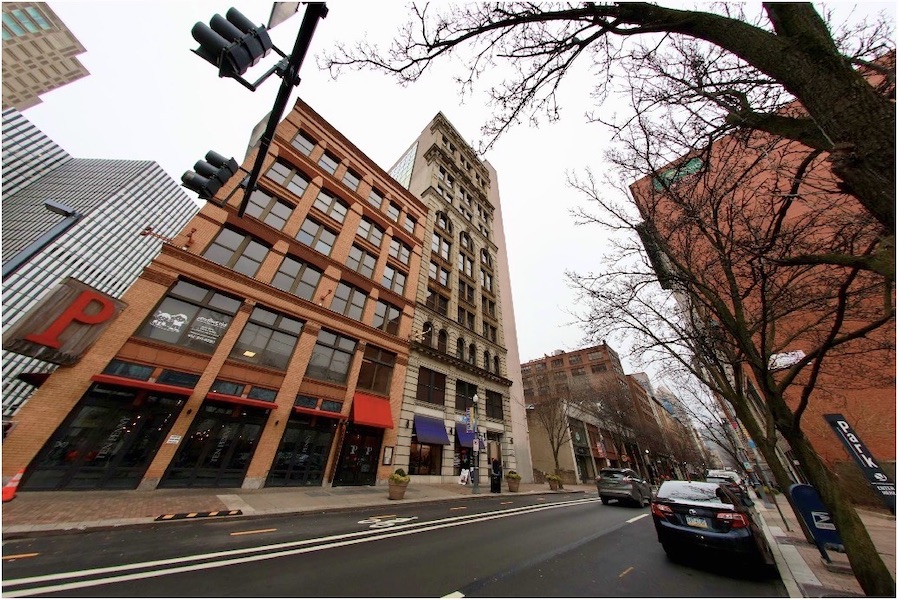}}
\subfloat[Manhattan Planes Rectified]{
\includegraphics[height=\h]{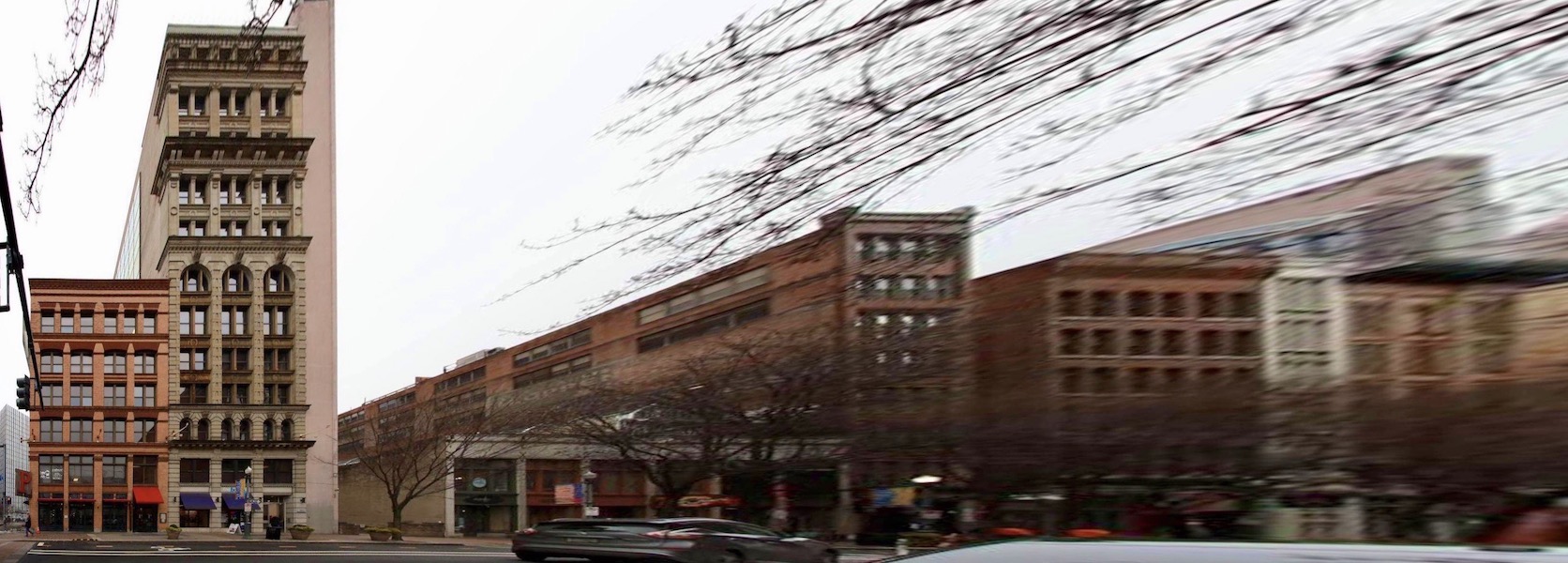}
\includegraphics[height=\h]{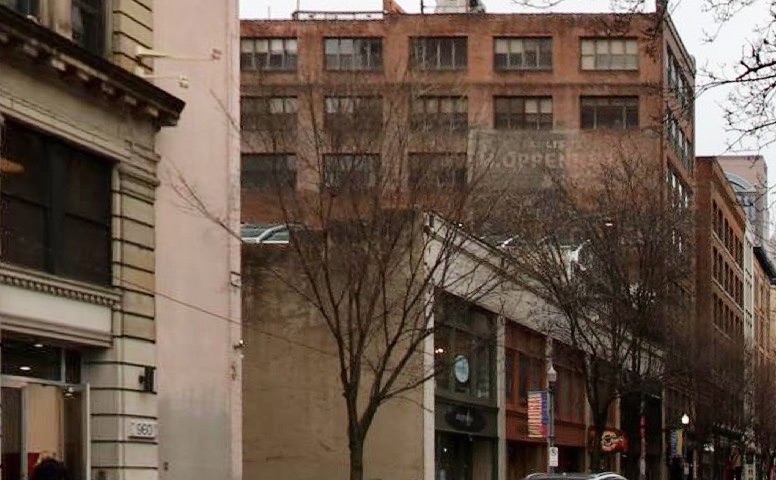}
\includegraphics[height=\h]{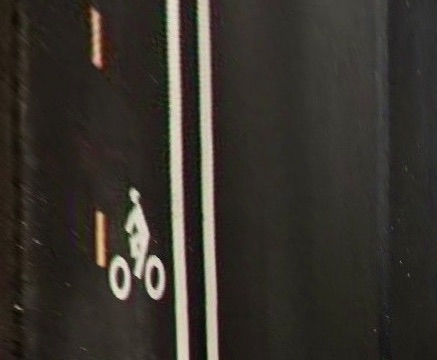}}
\vspace{-5pt}
\caption{(a) Input is an image. (b) Lens distortion corrected, and (c)
  Scene planes rectified. The method is fully automatic.}
\label{fig:first}
\end{figure}
}
\makeatother
\maketitle

\begin{abstract}
  This paper proposes minimal solvers that use combinations of imaged
  translational symmetries and parallel scene lines to jointly
  estimate lens undistortion with either affine rectification or
  focal length and absolute orientation. We use constraints provided
  by orthogonal scene planes to recover the focal length.  We show
  that solvers using feature combinations can recover more accurate
  calibrations than solvers using only one feature type on scenes that
  have a balance of lines and texture.  We also show that the proposed
  solvers are complementary and can be used together in a
  \RANSAC-based estimator to improve auto-calibration
  accuracy. State-of-the-art performance is demonstrated on a standard
  dataset of lens-distorted urban images. The code is available at
  \url{https://github.com/ylochman/single-view-autocalib}.
\end{abstract}

\section{Introduction}
\label{sec:introduction}
Imaged scene plane rectification and single-view camera
auto-calibration are closely related computer-vision tasks.  Both
tasks are ill-posed single-view geometry estimation problems that are
further complicated if the input image is distorted
\cite{Antunes-CVPR17,Pritts-BMVC16}.
In the presence of imaging noise, good feature coverage over large
parts of the image is necessary to observe the joint effects of
perspective warp and lens distortion. State-of-the-art techniques for
auto-calibrating or rectifying lens-distorted images use either
circular arcs fitted to edge detections or covariant region detections
as
inputs \cite{Antunes-CVPR17,Wildenauer-BMVC13,Pritts-IJCV20,Pritts-PAMI20}. Complementary
features can provide measurements on image regions that lack either
texture or lines. Each feature type---circular arc or covariant
region---has distinct advantages. Circular arcs provide accurate
measurements of imaged scene lines. However, it's difficult to group
arcs that are the images of parallel scene lines by appearance.  While
regions do not give the accuracy of circular arcs, discriminative
embeddings exist that can be used to cluster imaged coplanar
repeats \cite{Arandjelovic-CVPR12,Bay-ECCV06,Lowe-IJCV04,Mishchuk-NIPS17}.
Furthermore, multiple point correspondences can be extracted from one
region correspondence.
We propose solvers that combine the best of both worlds and
demonstrate that solvers using complementary feature types can extend
high-accuracy rectification and auto-calibration to challenging
highly-distorted images with diverse scene content
(see \figref{fig:first}).

\setlength\aboverulesep{0pt}
\setlength\belowrulesep{0pt}
\setlength\cellspacetoplimit{2pt}
\setlength\cellspacebottomlimit{2pt}

\begin{table*}[!t]
\centering
\begin{tabularx}{\textwidth}{Sl ScScS{Y} ScScS{Y}}
\toprule
& \multicolumn{3}{c}{Coplanar Configuration} & \multicolumn{3}{c}{Manhattan Configuration}\\[-5pt]
& \multicolumn{2}{c}{Inputs} & \multirow{2}{*}{Outputs} & \multicolumn{2}{c}{Inputs} & \multirow{2}{*}{Outputs} \\
& PC & CA &  & PC & CA &  \\
\midrule
\WildBMVC\cite{Wildenauer-BMVC13} &
0 & 
3+2 & 
\{1 VP, $\lambda$\} $\rightarrow$ 1 VP $\rightarrow$ $f$ &
0 & 
3+1+1 & 
\{1 VP, $\lambda$\} $\rightarrow$ \{2 VP, $f$\} \\
\AntunCVPR\cite{Antunes-CVPR17} &
0 & 
4+3 & 
2 VP $\rightarrow$ \{$\lambda$, $\ve[c]$\} $\rightarrow$ $f$ &
- &
- &
- \\
\EVP\cite{Pritts-CVPR18} &
2+2 &
0 & 
\{2 VP, $\lambda$, $\vl$\} $\rightarrow$ $f$ &
- &
- &
- \\
\rowcolor[gray]{0.95} \HybridSolver &
2+2 &
2 & 
\{3 VP, $\lambda$, $\vl$\} $\rightarrow$ $f$ &
3 &
1+1 & 
\{1 VP, $\lambda$\} $\rightarrow$ \{2 VP, $f$\} \\
\rowcolor[gray]{0.95} \HybridSolverTwo & 
2 &
2+2 &
\{3 VP, $\lambda$, $\vl$\} $\rightarrow$ $f$ &
2 &
2+2 & 
\{3 VP, $\lambda$, $f$\} \\
\rowcolor[gray]{0.95} \CircleDegSolver &
0 &
3+2 &
\{1 VP, $\lambda$\} $\rightarrow$ 1 VP $\rightarrow$ $f$ &
0 &
3+1+1 &
\{1 VP, $\lambda$\} $\rightarrow$ \{2 VP, $f$\} \\
\rowcolor[gray]{0.95} \CircleSolver &
0 &
2+2+2 &
\{3 VP, $\lambda$, $\vl$\} $\rightarrow$ $f$ &
0 &
2+2+2 &
\{3 VP, $\lambda$, $f$\} \\
\bottomrule
\end{tabularx}
\caption{Inputs and outputs of the state-of-the-art vs. the
proposed solvers (shaded in grey). The inputs are counted by the
number of corresponded features required for each estimated vanishing
point. Denotations are PC for point correspondence, CA for circular
arcs, VP for vanishing point, $\lambda$ for the division model
parameter, $\ve[c]$ for the distortion center, $\vl$ for the vanishing
line, and $f$ for the focal length. A sets of outputs is jointly
recovered by the solver, and right arrows indicate a chain of
estimates. The output at each step depends on the configuration of
VPs: either the VPs are coplanar, or they are oriented as a Manhattan
frame in the scene.}
\label{tab:state_of_the_art}
\vspace{-15pt}
\end{table*}

A solver whose inputs are sets of different geometric primitives is
called a \emph{hybrid solver} or \emph{mixed
solver} \cite{Camposeco-CVPR18}. Minimal solvers constructed from
points and lines were used in pose estimation \cite{Ramalingam-ICRA11,
Miraldo-ECCV18} and combinations of points and planes were explored in
SLAM \cite{Taguchi-ICRAr13}. The proposed hybrid solvers are the first
affine-rectifying and auto-calibrating solvers to admit complementary
geometric primitives---combinations of point correspondences provided
by a coplanar repeated region and circular arcs fitted to the linked
edge extractions of imaged parallel lines. In addition, we propose two
solvers whose inputs are circular arcs in a novel configuration, which
provides an additional feature sampling flexibility.

The affine-rectifying solvers can be adapted for metric rectification
and auto-calibration, which are the tasks evaluated in this paper.
The relative angle between the preimage of input features is assumed known, which provides
sufficient constraints for auto-calibration. Right angles are chosen
because they are the most common in man-made scenes. Three mutually
orthogonal directions in the scene are used to define a linear basis
called the \emph{Manhattan frame.} The presence of a Manhattan frame
in the scene is assumed by the auto-calibrating solvers.

All proposed solvers are derived from common constraints and are
related by a unified derivation. The derivation uses elementary
techniques from projective geometry, which makes solver generation
straightforward. The solvers are fast, stable and robust, which makes
them good candidates to be used in \RANSAC-based
estimators \cite{Fischler-CACM81}. 

The solvers are also used as an ensemble in a \RANSAC estimator that
samples combinations of arcs and point correspondences according to
the required input of the invoked solver. Solvers are invoked with
different frequencies depending on the content of the scene. The use
of multiple solvers that have different input types is similar to the
hybrid \RANSAC approach introduced
in \cite{Camposeco-CVPR18,Mateus-CVPR20}. We show that sampling
combinations of regions and arcs improves auto-calibration accuracy
over using either only regions or arcs.

\section{Related Work}
\label{sec:related_work}
The state of the art has extended scene-plane rectification and
single-view auto-calibration to lens-distorted images. The common
choice of parameterization for lens undistortion is the division
model. It is preferred in ill-posed settings since it has only one
parameter and can effectively model a wide range of radial lens
undistortions \cite{Fitzgibbon-CVPR01}.

Antunes \etal \cite{Antunes-CVPR17} and
Wildenauer \etal \cite{Wildenauer-BMVC13} are two methods that recover
vanishing points from distorted images by using the constraint that
parallel scene lines are imaged as circles intersecting at their
distorted vanishing point under the division
model \cite{Bukhari-JMIV13,Fitzgibbon-CVPR01,Strand-BMVC05,Wang-JMIV09}. The same constraint is used for the proposed solvers.
Sets of circular arcs whose preimages are parallel lines are used to induce
constraints on the division-model parameter and vanishing
point. Vanishing points are recovered, and auto-calibration is
estimated by assuming that vanishing points correspond to imaged
Manhattan frame directions. Antunes \etal
\cite{Antunes-CVPR17} require a set of four circular arcs for
the first vanishing point and three for the second, while Wildenauer
\etal \cite{Wildenauer-BMVC13} requires three for the first and two for the second or three for the first, one for the second, and one for the third.

Pritts \etal \cite{Pritts-CVPR18,Pritts-ACCV18,Pritts-IJCV20,Pritts-PAMI20}
proposed a suite of solvers that can jointly undistort and
affinely-rectify from the imaged translation directions of coplanar
repeated scene texture. The solvers directly estimate the vanishing
line of the scene plane but also return the vanishing directions of
the imaged translations that are consistent with the recovered
vanishing line. The solver variant of
\cite{Pritts-CVPR18,Pritts-PAMI20} requiring two correspondences of
covariant regions can be directly used for auto-calibration, if the
recovered vanishing points are imaged Manhattan frame directions.

See \tabref{tab:state_of_the_art} for a comparison of the proposed
solvers to the state of the art. The solver notation refers the input
configurations, e.g., \textbf{4PC} means four point correspondences, while \textbf{6CA} means six circular arcs are required.

\section{Preliminaries}
\label{sec:preliminaries}
We assume the division model \cite{Fitzgibbon-CVPR01} of lens
undistortion
\begin{equation}
\label{eq:division_model}
\begin{split}
g(\vxd,\lambda) = \rowvec{3}{\xd}{\yd}{1+\lambda\rd^2}^{\T},
\end{split}
\end{equation}
where $\lambda$ encodes the magnitude and type (by sign) of radial distortion, $\vxd=\rowvec{3}{\xd}{\yd}{1}^{\T}$ is a homogeneous image point with radius $\rd^2 = \xd^2+\yd^2$, and the origin is at the distortion center.
We assume an orthogonal raster with unit aspect ratio and fix the distortion center and the principal point at the image center. In this setting the camera's intrinsic matrix is $\mK = \diag{f,f,1}$, where $f$ is the focal length in pixels. The projection of a homogeneous scene point $\vX=\rowvec{4}{X}{Y}{Z}{1}^{\T}$ onto \vxd is as following 
\begin{equation}
\label{eq:projection}
\begin{split}
\gamma g(\vxd,\lambda) = \mK 
\begin{bmatrix}\mR ~|~ \ve[t]\end{bmatrix}
\vX,
\end{split}
\end{equation}
where $\mR $ is a $3\times3$ rotation matrix, $\ve[t]$ is a translation vector.

\section{Minimal Solvers}
\label{sec:minimal_solvers}

\begin{figure}[!t]
\centering
\captionsetup[subfigure]{}
\subfloat[Coplanar Configuration]
{\includegraphics[height=0.23\columnwidth]{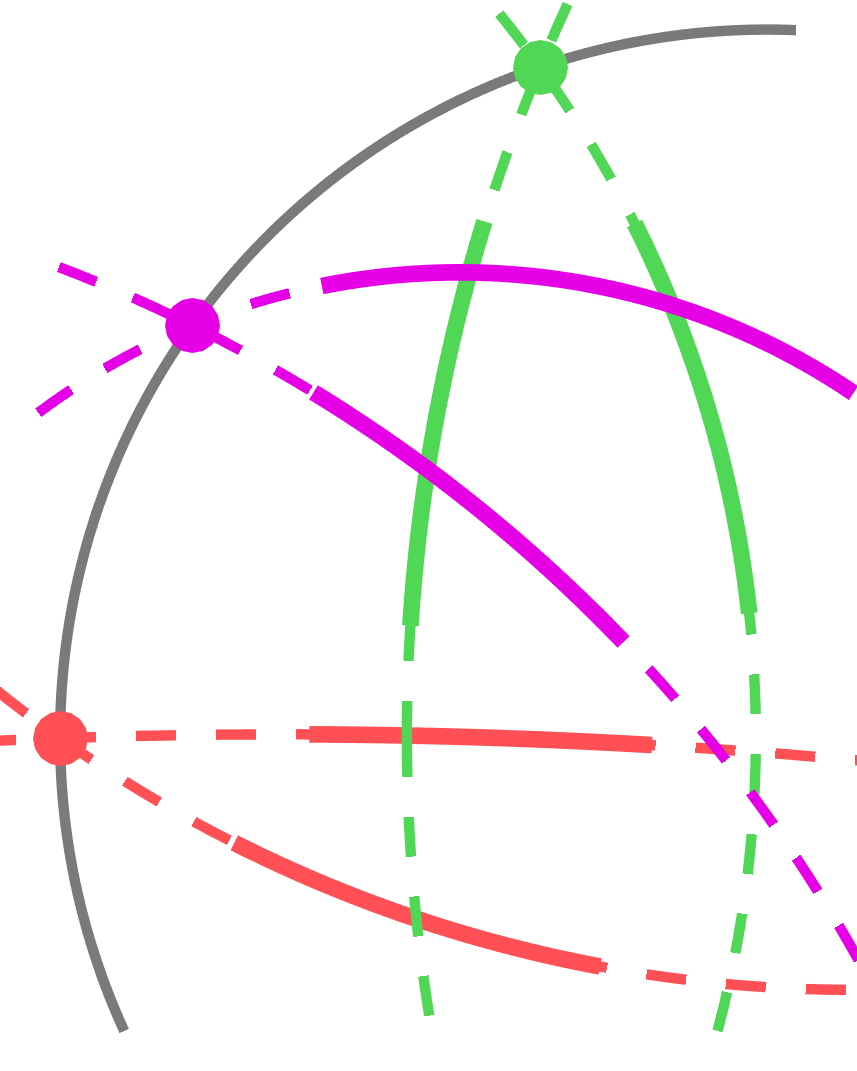}~
\includegraphics[height=0.23\columnwidth]{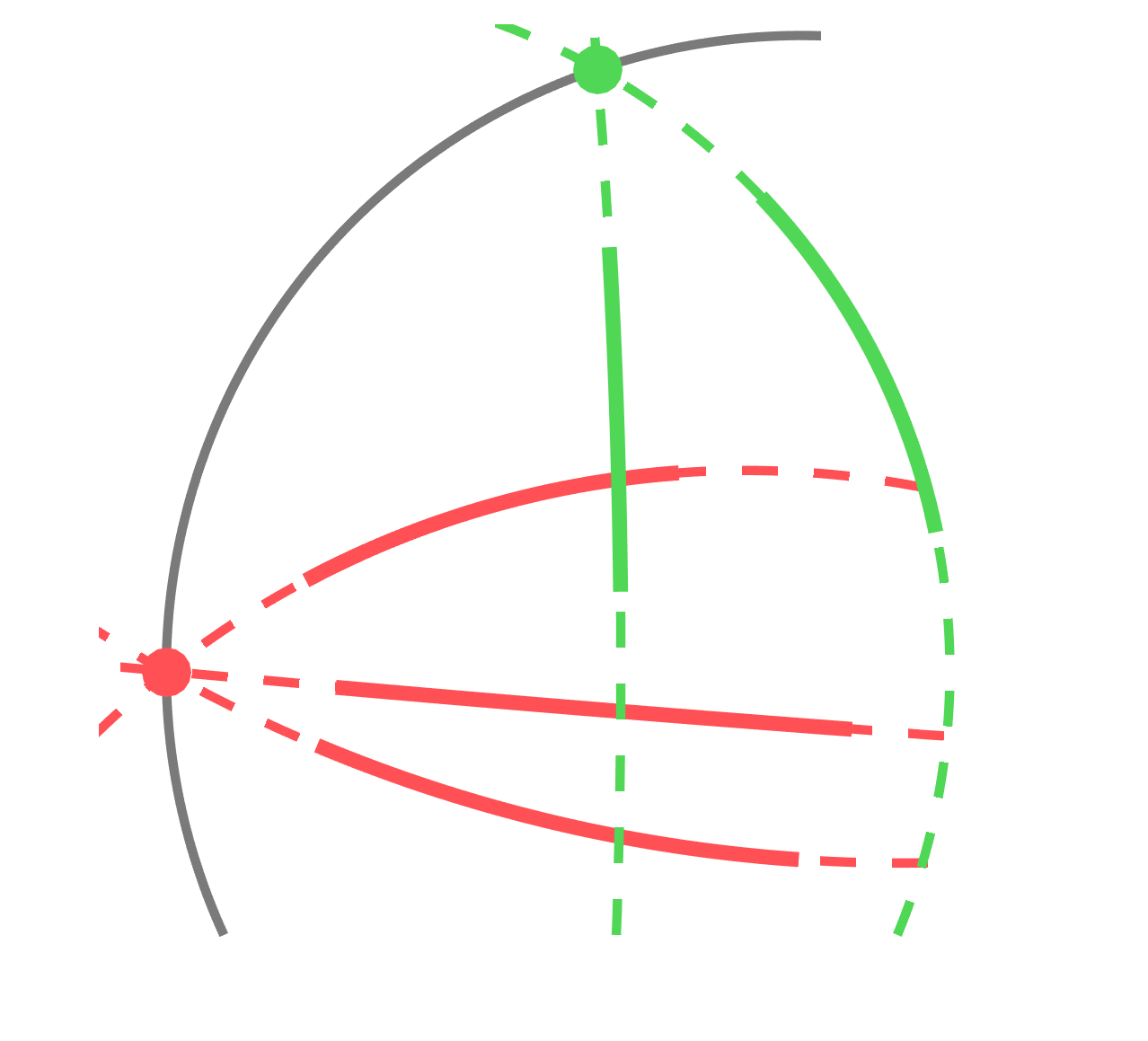}}~
\subfloat[Manhattan Configuration]
{\includegraphics[height=0.23\columnwidth]{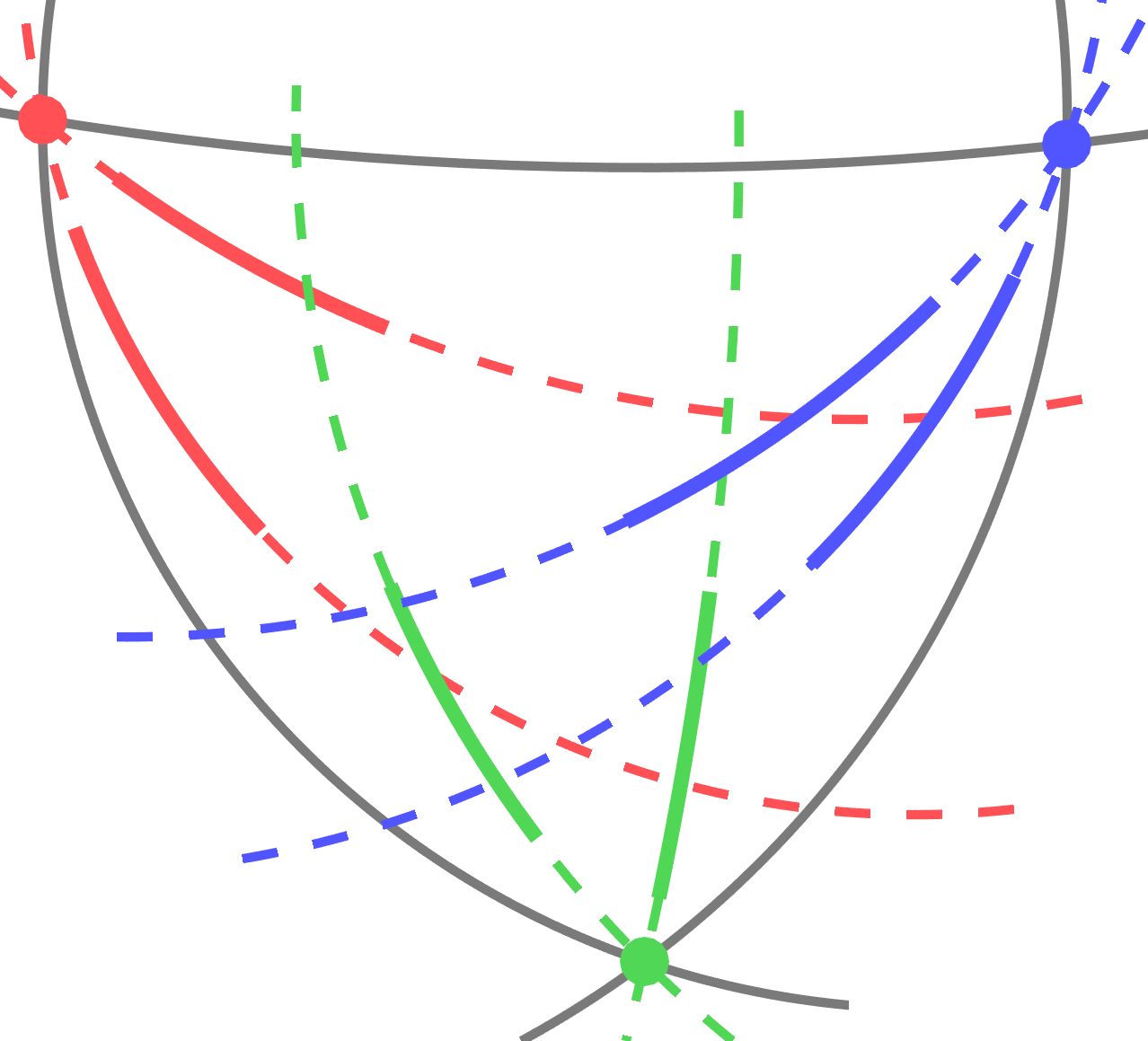}~
\includegraphics[height=0.23\columnwidth]{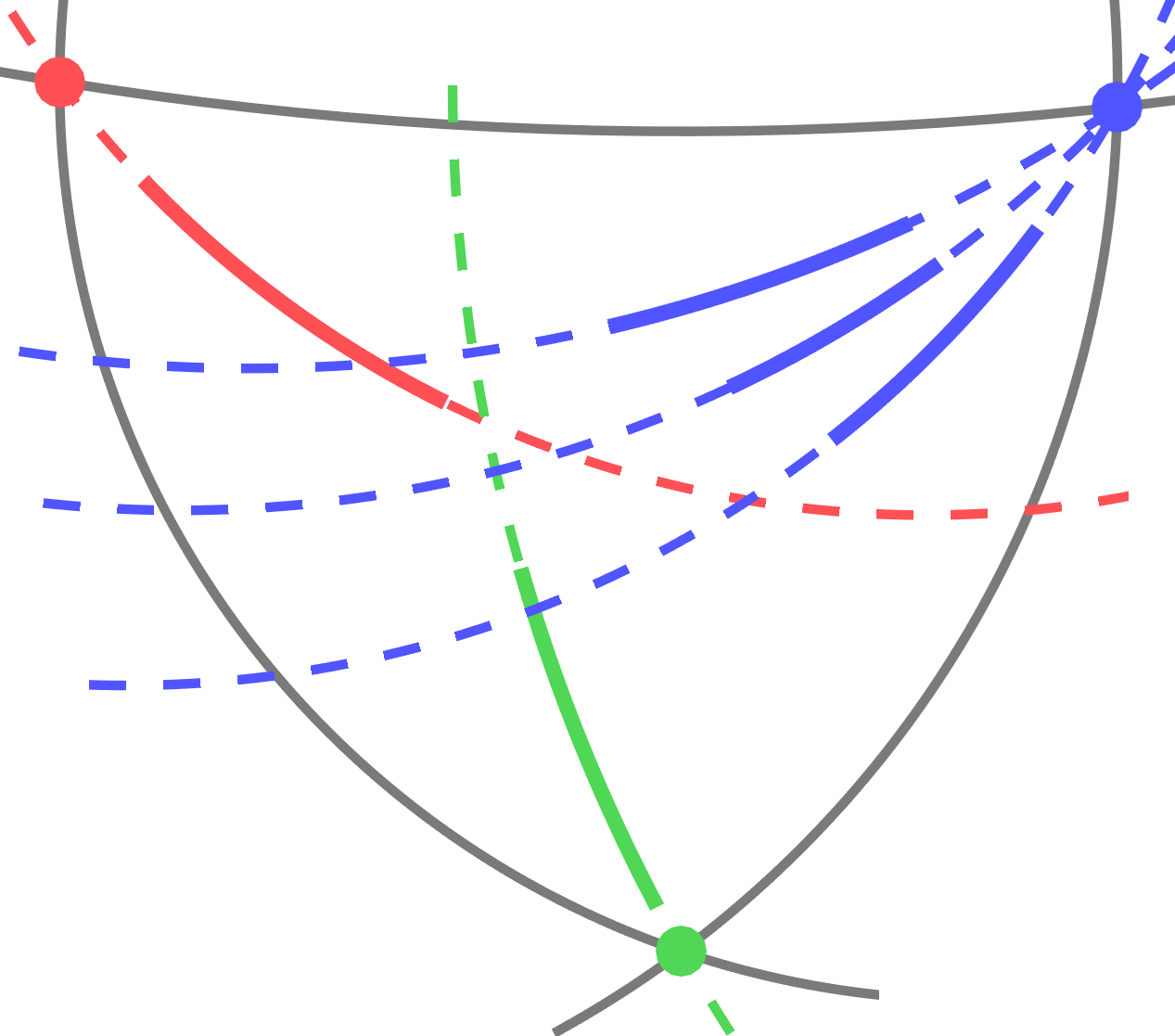}}
\caption{Input configurations of the proposed solvers. The vanishing points are constructed from either circular arcs or point correspondences. Inputs are color coded with respect to their corresponding vanishing points that are (a) coplanar (see \secref{sec:rect_solvers}) or (b) mutually orthogonal (see \secref{sec:ac_solvers}).}
\label{fig:solvers_inputs}
\vspace{-15pt}
\end{figure}
All the proposed solvers use the invariant
that imaged parallel scene plane lines intersect at the vanishing
point of their imaged translation direction.  A vanishing point
$\vu(\lambda)$ is constructed as the meet of the undistorted images
$\vm(\lambda)$ and $\vmp(\lambda)$ of parallel lines,
\begin{equation}
  \label{eq:meet} \vu(\lambda)
  = \vm(\lambda) \times \vmp(\lambda).
\end{equation}
The undistorted images of parallel lines can be constructed from
distorted measurements under \eqref{eq:division_model}. There are two
different parametric forms of $\vm(\lambda)$ and $\vmp(\lambda)$
depending on the type of measurements used: either circular arcs or
point correspondences.

The relative orientation of the detected vanishing points in not known
apriori. The solvers handle two cases: either the vanishing points are
coplanar, or they are the image of a Manhattan frame (see
\figref{fig:solvers_inputs}).

\subsection{Coplanar Vanishing Points}
\label{sec:rect_solvers}
The joint rectifying solvers share a common derivation based on the
invariant that vanishing points are incident to the scene plane's
vanishing line. The vanishing point-vanishing line incidence equation
is $\vu^{\T}\vl=0$. There are four unknowns to be recovered, namely
$\vl=\rowvec{3}{l_1}{l_2}{l_3}^{\T}$ and the division model parameter
$\lambda$. The vanishing line \vl is homogeneous, so it has only two
degrees of freedom. Thus, three scalar constraint equations of the
form $\vu[i](\lambda)^{\T}\vl=0$ are needed, where
\buildset{\vu[i]}{i=1}{3} are three distinct vanishing points
constructed as in \eqref{eq:meet}.
With a matrix $\ma{U}(\lambda)$ formed by the vanishing points $\{\vu[i]\}$, the point-line incidence constraints can be concisely written as a
homogeneous matrix-vector equation,
\begin{equation}
\begin{split}
\ma{U}(\lambda)\vl =
\begin{bmatrix}
  \left(\vm[1](\lambda) \times \vmp[1](\lambda)\right)^{\T} \\
  \left(\vm[2](\lambda) \times \vmp[2](\lambda)\right)^{\T}  \\
  \left(\vm[3](\lambda) \times \vmp[3](\lambda)\right)^{\T}  
\end{bmatrix}\vl
=
\ve[0].
\label{eq:stacked_constraints}
\end{split}
\end{equation}
The system of equations \eqref{eq:stacked_constraints} has
a non-trivial solution \vl only if $\ma{U}(\lambda)$ is singular,
which generates the scalar constraint equation $\det \ma{U}(\lambda) =
0$ on $\lambda$.
The parameterization for lines
$\vm[i](\lambda)$ and $\vmp[i](\lambda)$ used by all solver variants
results in a quartic equation in $\lambda$, which can be solved in
closed form. After recovering $\lambda$, the null space of $\ma{U}$ is
computed, which gives the vanishing line \vl. The system of equations
in \eqref{eq:stacked_constraints} is agnostic to the construction
method for the \vu[i], which gives a unified way to generate
rectifying solvers that use different feature types.

\begin{figure}[!t]
\centering
\setlength{\h}{0.48\columnwidth}
\captionsetup[subfigure]{}
\begin{minipage}{0.49\columnwidth}
\subfloat[VP Labeling]{
\includegraphics[width=\h]{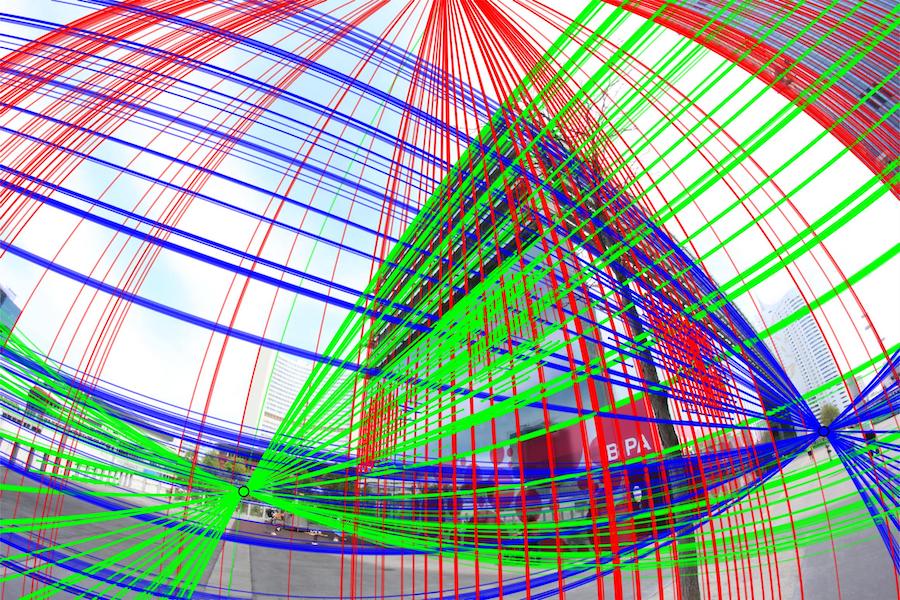}}
\end{minipage}
\begin{minipage}{0.49\columnwidth}
\subfloat[VL Labeling]{
\includegraphics[width=\h]{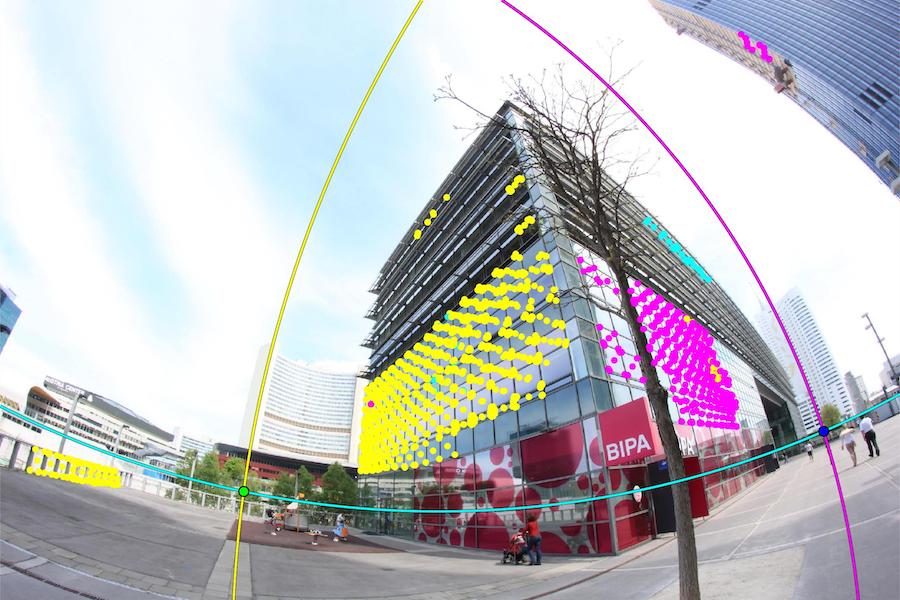}}
\end{minipage}
\caption{(a) The distorted vanishing points of the imaged Manhattan frame are colored RGB. Circular arcs and region correspondences are colored w.r.t. their assigned VPs. (b) Imaged vanishing lines (VL) are colored CMY Covariant regions are colored w.r.t. their assigned VLs.}
\vspace{-15pt}
\label{fig:detected_manhattan_frame}
\end{figure}

\newcommand{\addrowss}[1]{
  \includegraphics[height=0.12\textwidth]{suppimg/#1/#1.jpg} &
  \includegraphics[height=0.12\textwidth]{suppimg/#1/#1__ud__CircleSolver+HybridSolverTwo.jpg} &
  \includegraphics[width=0.19\textwidth]{suppimg/#1/#1__rect1__CircleSolver+HybridSolverTwo.jpg} &
  \includegraphics[width=0.19\textwidth]{suppimg/#1/#1__rect2__CircleSolver+HybridSolverTwo.jpg} &
  \includegraphics[height=0.12\textwidth]{suppimg/#1/#1__rect3__CircleSolver+HybridSolverTwo.jpg}\\
}

\renewcommand{\tabcolsep}{2pt}
\newcolumntype{P}[1]{>{\centering\arraybackslash}p{#1}}
\begin{figure*}[!t]
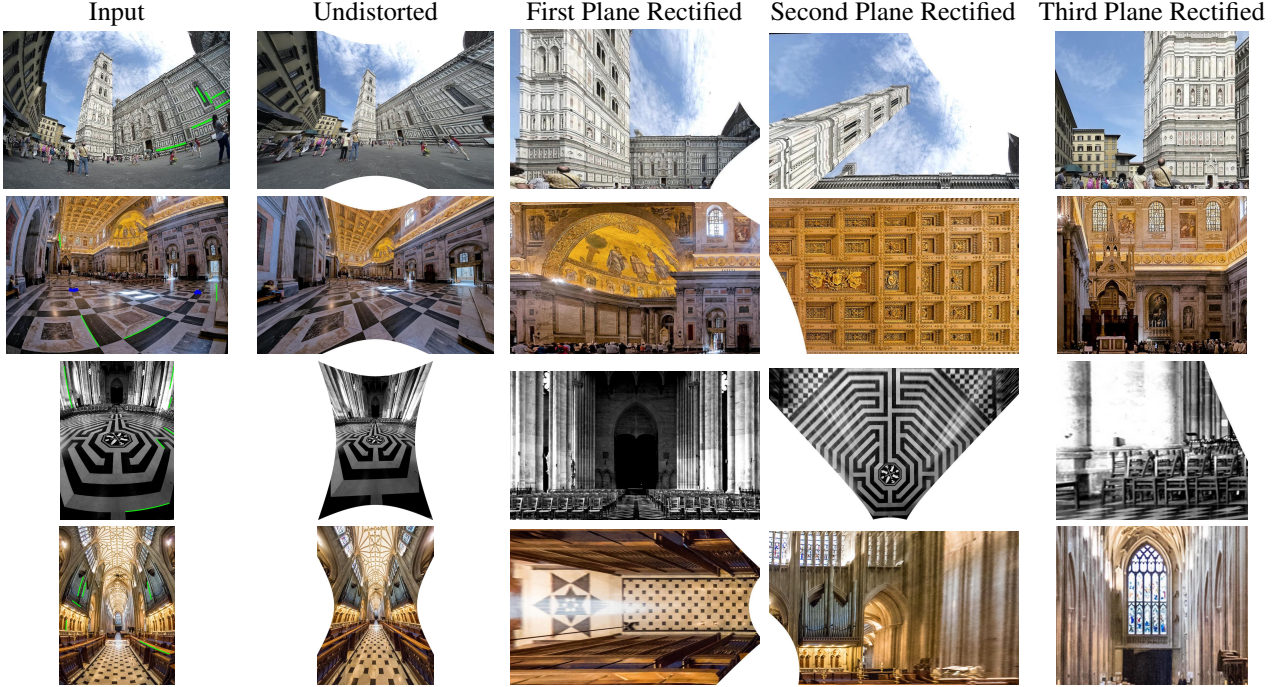

\centering
\captionsetup[subfigure]{labelformat=empty}
{\def\arraystretch{0.7}
\begin{tabular}{P{3.3cm} P{3.3cm} P{3.3cm} P{3.3cm} P{3.3cm}}
Input & Undistorted & First Plane Rectified & Second Plane Rectified & Third Plane Rectified\\[1pt]
\addrowss{13}\\[-8pt]
\addrowss{37}\\[-8pt]
\addrowss{41}\\[-8pt]
\addrowss{42}
\end{tabular}
}
\caption{Auto-calibration results on wide-angle imagery. The scene planes oriented with the Manhattan frame are rectified. The minimal sample ---green circles and blue regions---of the returned solution is depicted on the input image.}
\vspace{-15pt}
\label{fig:real_data_internet}
\end{figure*}

\vspace{-10pt}
\paragraph{Auto-Calibration Upgrade}
The Manhattan frame directions defined by the standard basis
vectors $\buildset{\ve[e]_i}{i=1}{3}$
are
imaged to the distorted vanishing points $\buildset{\ve[\tilde{u}]_i}{i=1}{3}$ by \eqref{eq:projection}
\begin{equation}
\ve[u]_i = \gamma g(\ve[\tilde{u}]_i,\lambda) = \mK\mR\ve[e]_i, \quad  i \in \buildset{1 \ldots 3}{}{}.
\end{equation}
See an example of detected imaged Manhattan frame in \figref{fig:detected_manhattan_frame}. Two finite vanishing points $\ve[u]_1=\rowvec{3}{u_{x1}}{u_{y1}}{u_{w1}}^{\T}$,
$\ve[u]_2=\rowvec{3}{u_{x2}}{u_{y2}}{u_{w2}}^{\T}$ that are undistorted images of two Manhattan frame directions are sufficient to recover the focal length~\cite{Wildenauer-CVPR12}
\begin{equation}
  \label{eq:autocalib_f}
  f = \sqrt{\dfrac{u_{x1}u_{x2}+u_{y1}u_{y2}}{-u_{w1}u_{w2}}}.
\end{equation}
The relative orientation of the camera with respect to the Manhattan frame is then computed as following
\begin{equation}
\label{eq:autocalib_R} 
\mR =
\begin{bmatrix}\strut
\dfrac{\mK^{\inv}\vu[1] \strut}{\strut \| \mK^{\inv}\vu[1]\|} & 
\dfrac{\mK^{\inv}\vu[2] \strut}{\strut \| \mK^{\inv}\vu[2]\|} &
\dfrac{\mK^{\T} (\vu[1] \times \vu[2]) \strut} {\strut \| \mK^{\T} (\vu[1] \times \vu[2]) \|}\strut
\end{bmatrix}.
\end{equation}
A conjugate rotation $\mK\mR^\T\mK^{-1}$ is used to metrically rectify the Manhattan planes (see \figsref{fig:first}, \ref{fig:real_data_internet}, and \ref{fig:real_data_AIT}).

\subsection{Orthogonal Vanishing Points}
\label{sec:ac_solvers}
The joint auto-calibrating solvers use an invariant that the preimages
of the vanishing points are mutually orthogonal scene directions. The
orthogonality constraint is written as $\vu[i]^\T \omega \vu[j] = 0$,
where $\omega = \text{diag}(1/f^2,1/f^2,1)$ is the image of an
absolute conic. The functional forms of the vanishing points
$\vu(\lambda)$ are constructed by \eqref{eq:meet}. There are two
unknowns to be recovered --- $\lambda$ and $f$. Thus, two scalar
constraint equations of the form $\vu[i]^\T \omega \vu[j] = 0$ are
needed, which means that three vanishing points
$\buildset{\vu[i]}{i=1}{3}$ are needed. Rewriting the constraint
equations in the matrix form gives
\begin{equation}
\label{eq:ac_stacked_constraints}
\begin{bmatrix}
\vu[1](\lambda)\odot\vu[2](\lambda)\\
\vu[1](\lambda)\odot\vu[3](\lambda)\\
\vu[2](\lambda)\odot\vu[3](\lambda)\\
\end{bmatrix}
\colvec{3}{1}{1}{f^2} = \ve[0],
\end{equation}
where $\odot$ is the entrywise product. By substituting $f$ from the first row into the second and the third rows we obtain the system of equations only in $\lambda$.
The coordinates of a vanishing point constructed from circular arcs or from point correspondences are either linear or quadratic in
$\lambda$, which gives two polynomial equations in $\lambda$ of order
six. We solve the
resulting system for $\lambda$, recover $f$ from the linear system
\eqref{eq:ac_stacked_constraints}, and compute \mR by \eqref{eq:autocalib_R}. Similarly to \eqref{eq:stacked_constraints}, the system of equations in \eqref{eq:ac_stacked_constraints} is
agnostic to the construction method for the \vu[i].

\begin{figure*}[!t]
\centering
\setlength\fwidth{0.55\columnwidth}
\subfloat[Numerical Stability]{
\scriptsize{\input{fig/stability__warperr.tikz}}
\label{fig:stability}}
\hfill
\subfloat[Sensitivity: \textbf{PC} and \textbf{PC+CA} Solvers]{
\scriptsize{\input{fig/sensitivity_region_based__warperr.tikz}}
\label{fig:sensitivity_rgns}}
\hfill
\subfloat[Sensitivity: \textbf{CA} Solvers]{
\scriptsize{\input{fig/sensitivity_circle_based__warperr.tikz}}
\label{fig:sensitivity_circles}}
\\[-2pt]
\centering
\small{\begin{tikzpicture}
\begin{customlegend}
[legend columns=-1,
legend style=
{draw=none,/tikz/every even column/.append style={column sep=0.25cm},cells={align=center}},
legend cell align={left},
legend entries={
\WildBMVC~\cite{Wildenauer-BMVC13},
\EVP~\cite{Pritts-CVPR18},
~,
\HybridSolver,
\HybridSolverTwo,
\CircleDegSolver,
\CircleSolver,
}
]
\addlegendimage{WildBMVC,fill=WildBMVC,only marks,mark=square*}
\addlegendimage{EVP,fill=EVP,only marks,mark=square*}
\addlegendimage{white,fill=white,only marks,mark=circle*}
\addlegendimage{HybridSolver,fill=HybridSolver,only marks,mark=square*}
\addlegendimage{HybridSolverTwo,fill=HybridSolverTwo,only marks,mark=square*}
\addlegendimage{CircleDegSolver,fill=CircleDegSolver,only marks,mark=square*}
\addlegendimage{CircleSolver,fill=CircleSolver,only marks,mark=square*}
\end{customlegend}  
\end{tikzpicture}}\\
\vspace{-7pt}
\mycaption{Numerical Stability and Noise Sensitivity of Solvers}{
(a) Histogram of the $\log_{10}$ warp error for 1000 synthetic scenes with noiseless features.
(b-c) RMS warp error \rmswarperr after 25 iterations of a simple \RANSAC on 1000 synthetic scenes with increasing levels of noise $\sigma$ added to the point correspondences and/or circular arcs. Results are shown for (b) the region based solver \EVP \cite{Pritts-CVPR18, Pritts-PAMI20} and proposed hybrid solvers \HybridSolver and \HybridSolverTwo; (c) the arc-based solvers \WildBMVC \cite{Wildenauer-BMVC13}, and proposed \CircleDegSolver and \CircleSolver.}
\label{fig:stability_and_sensitivity}
\vspace{-15pt}
\end{figure*}

\subsection{Coincident Vanishing Points}
The system of equations \eqref{eq:stacked_constraints} is
trivially singular if two of the vanishing points from \buildset{\vu[i]}{i=1}{3} are coincident.
This occurs if two vanishing points are constructed by drawing two pairs of imaged lines that are mutually parallel in the scene. Similarly, the orthogonality constraint equations \eqref{eq:ac_stacked_constraints} are inconsistent if two of the vanishing points from \buildset{\vu[i]}{i=1}{3} are coincident. Vanishing
point coincidence can be used to place a constraint on the
division model parameter
\begin{equation}
\label{eq:coincident_vp}
\vu[i](\lambda) \times \vu[j](\lambda) = \ve[0].
\end{equation}
The coordinates of a vanishing point are either linear or quadratic in
$\lambda$, which gives two cubic equations and one quadratic equation
in $\lambda$ in the system of \eqref{eq:coincident_vp}. After solving
for $\lambda$ we back-substitute and recover $\vl$ from
\eqref{eq:stacked_constraints}, if the vanishing points are collinear,
or solve for $f$ and $\ma{R}$ using \eqref{eq:ac_stacked_constraints},
if they are the image of a Manhattan frame.

\subsection{Solver Generation}
\label{sec:solver_generation}
The solvers can be generated by constructing vanishing points
$\{\vu[i]\}$ from circular arcs or vanishing points $\{\vu[j]\}$ from
imaged translated point correspondences, where $i \neq j$ and $i,j \in
1 \ldots 3$.  We construct three solver
variants: \begin{enumerate*}[(i)] \item the hybrid solver \HybridSolver,
  that uses a pair of circular arcs and two pairs of imaged translated
  point correspondences, \item the hybrid solver \HybridSolverTwo
  admitting two pairs of circular arcs and a pair of imaged translated
  point correspondences, \item and the solver \CircleSolver that uses
  three pairs of circular arcs.\end{enumerate*} It is possible to construct a solver from only point
correspondences. If coplanar vanishing points are assumed, then the
solver is the same as the \EVL solver proposed in
\cite{Pritts-PAMI20}. However, covariant region detections typically used to
extract point correspondences are unlikely to provide orthogonal structures \cite{Pritts-PAMI20}. The case where two constructed vanishing points are coincident (see
\eqref{eq:coincident_vp}) is handled by the fourth proposed \CircleDegSolver
solver. Three parallel scene lines are sufficient to provide
constraints on two vanishing points that coincide. Thus the
\CircleDegSolver requires a triple and a pair of circular arcs. Inputs of the solvers are listed in \tabref{tab:state_of_the_art} and input
configurations are shown in \figref{fig:solvers_inputs}.

\figref{fig:real_data_internet} shows the qualitative performance of
the solvers on challenging images from the Internet.
See also \secref{sec:supp_real_data} in Supplemental for more results.

\subsection{Line Construction}
\label{sec:line_construction}
The solvers use point correspondences and sets of circular arcs, where
a set of arcs tentatively consists of the images of parallel scene
lines under the division model. Point correspondences consist of
points extracted from the image of translational symmetries. The
preimages of the point correspondences must have the same translation
direction and distance of translation in the scene plane. To reduce
the expected number of \RANSAC samples, we extract point
correspondences from covariant region correspondences.

\vspace{-10pt}
\paragraph{Imaged Translated Coplanar Repeats}
The corresponding points of two coplanar translated repeated regions
\cite{Schaffalitzky-BMVC98} form parallel scene lines. Thus we can use
the point correspondences extracted from distorted images of coplanar
repeated texture \cite{Pritts-CVPR18,Pritts-IJCV20} to construct the
undistorted images of parallel lines. Let \vxd and \vxdp be two
distorted points in correspondence. Then a line is constructed as a
join of their undistorted points,
\begin{equation}
  \label{eq:ct_meet} 
  \ve[t](\lambda) = g(\vxd,\lambda) \times g(\vxdp,\lambda).
\end{equation}
Using \eqref{eq:ct_meet}, parallel lines $\ve[t]_i(\lambda)$ and
$\ve[t]^{\prime}_i(\lambda)$ can be constructed from points extracted
from imaged translational symmetries to provide the constraints
required by \eqref{eq:stacked_constraints} to recover $\lambda$ and
\vl. The coordinates of vanishing point
$\vu[i](\lambda)=\ve[t]_i(\lambda) \times \ve[t]^{\prime}_i(\lambda)$
are either linear or quadratic in $\lambda$.

\begin{table*}[!t]
\centering
\begin{tabularx}{\textwidth}{Sl *2{S{Y}} *4{>{\columncolor[gray]{0.95}}Sc} *4{>{\columncolor[gray]{0.75}}Sc}}
\toprule
Solver & 
\EVP &
\WildBMVC &
\HybridSolver &
\HybridSolverTwo &
\CircleDegSolver &
\CircleSolver &
Hybrid &
Arcs &
All &
\CircHyb \\
\midrule 
\% of Top-1  & 1.5\%~ &10.2\%~&15.5\%~&21.7\%~&25.4\%~&25.7\% & - & - & - & - \\
\midrule 
Median $\lambda$ Rel. Err. &
14.91 & 2.27 & 2.81 & 2.38 & 2.24 & 2.29 & 2.33 & 2.27 & \textbf{2.14} & 2.16\\
Median $f$ Rel. Err. &
25.93 & 1.42 & 1.82 & 1.48 & 1.42 & 1.4 & 1.48 & 1.39 & \textbf{1.38} & 1.39\\
Median \rmswarperr &
186.51 & 14.02 & 15.86 & 14.9 & 14.12 & 13.73 & 14.77 & 14.29 & 13.91 & \textbf{13.35}\\
\bottomrule
\end{tabularx}
\mycaption{Performance on AIT Dataset}{The Top-1 solution is the
  calibration that has the lowest RMS warp error among the
  solvers. The proposed solvers are in grey. Solver combinations used
  in Hybrid \RANSAC are in dark grey. \say{Hybrid} uses only the
  hybrid solvers \ie \HybridSolver and \HybridSolverTwo, \say{Arcs}
  uses only the arc solvers \ie \WildBMVC, \CircleDegSolver, and
  \CircleSolver, and \say{All} uses every solver. The AIT dataset was
  run ten times for a total of 1020 calibrations for each solver or
  solver combination.}
\label{tab:AIT}
\vspace{-15pt}
\end{table*}

\vspace{-10pt}
\paragraph{Imaged Scene Lines}
Lines are distorted to circles under the division
model \cite{Bukhari-JMIV13,Fitzgibbon-CVPR01,Strand-BMVC05,Wang-JMIV09},
and
the normals of a circle are mapped to the normals of the circle’s
undistorted image by the transposed inverse of the Jacobian of the
division model. This gives the following form of the undistorted line
$\ve[s](\lambda)$
\begin{equation}
  \label{eq:ud_tangent_line} \ve[s](\lambda) = 
  \lambda \colvec{3}{\tilde n_x \xd^2+2\tilde n_y \xd\yd-\tilde
  n_x\yd^2}{\tilde n_y \yd^2+2\tilde n_x\xd\yd- \tilde n_y \xd^2}{0}
  +\begin{pmatrix} \tilde
    n_x \\ \tilde n_y \\
    -\eve[\tilde{n}]^\T\eve[\tilde{x}] \end{pmatrix}\\
  ,
\end{equation}
where $\eve[\tilde{n}] =
\rowvec{2}{\tilde n_x}{\tilde n_y }^\T$ is a normal to a circle at the point $\eve[\tilde{x}]=\rowvec{2}{\xd}{\yd}^{\T}$ \cite{Wildenauer-BMVC13}. Thus circles that are distorted images of parallel scene lines can generate pairs of undistorted lines $\ve[s]_i(\lambda)$ and $\ve[s]_i^{\prime}(\lambda)$ that provide the constraints required by
\eqref{eq:stacked_constraints} to recover $\lambda$ and \vl. The
coordinates of vanishing point
$\vu[i](\lambda)=\ve[s]_i(\lambda)
\times \ve[s]^{\prime}_i(\lambda)$ are either linear or quadratic in
$\lambda$.

\section{Features}
\label{sec:feature_extraction}
Contours are constructed by linking sub-pixel Canny edge detections
with morphological operations. The contours are decimated by the
Ramer---Douglas---Peucker
algorithm \cite{Douglas-Cartographica73}. The maximum likelihood fit
of a circle to the contour is estimated by nonlinear least squares,
which is initialized by Taubin's bias-renormalization fit for conics
\cite{Taubin-PAMI91}.

Point correspondences are extracted from a covariant region
correspondence \cite{Perdoch-ICPR06,Pritts-IJCV20}.  In particular, we
use the Maximally-Stable Extremal Region detector with the local
affine-frame upgrade
\cite{Vedaldi-SOFTWARE08,Matas-BMVC02,Matas-ICPR02,Obdrzalek-BMVC02},
which provides three point correspondences.  Affine-covariant regions are tentatively labeled as coplanar repeated texture if the regions
are similar in appearance. Region appearance is embedded by the
RootSIFT descriptor and clustered into tentative repeats similar to
what was done in \cite{Pritts-PAMI20}.

\subsection{Rejecting Minimal Input Configurations}
\label{sec:consistency_measure}
The configuration of the input sample is apriori unknown. The possible
configurations are illustrated in \figref{fig:solvers_inputs}. Implausible
solutions from the invoked solvers that cannot handle the sampled
input configuration are rejected by testing their geometric
consistency with the minimal samples. Note that the minimal solution
will exactly satisfy the algebraic constraints of the solver with the
minimal sample as input. However, we test against additional
properties of the minimal sample that are unused by the
solver. Verification against the minimal sample incurs a negligible
and justifiable computation cost since it can prevent unneeded
consensus set evaluation by \RANSAC, which is an expensive
computation \cite{Matas-ICCV05}. The rejection test is outlined for
circular arcs and coplanar repeats in the next two paragraphs.

\vspace{-10pt}
\paragraph{Consistency of Circular Arcs}
\label{sec:circular_arc_consistency}
The minimal solution and the midpoint where the contour normal is
measured are used to construct a circle through the distorted
image of the vanishing point of the direction of the scene line
that generates the contour. This construction is similar to
a line through the vanishing point construction in the undistorted space proposed by Tardif
in \cite{Tardif-ICCV09}. The contour midpoint is undistorted to
$\bar{\vx}$, and its join with the vanishing point $\vm
= \bar{\vx} \times \vu$ is distorted to the circle
$\vmd=(a,b,c)$, where $(a,b)$ is the circle center and $c$ is the
radius.

The consistency measure is the mean squared orthogonal distance between points of the contour $\buildset{ \vxd[k] }{k=1}{K}$
and the circle $\vmd$ through its midpoint
\begin{equation}
\label{eq:arc_vp_consistency}
\mathcal{J} = \frac{1}{K} \sum_{k=1}^K (\sqrt{(\xd[k]-a)^2+(\yd[k]-b)^2} - c)^2.
\end{equation}
The construction is shown in \figref{fig:supp_circle_through_vp} of
the Supplemental.

\vspace{-10pt}
\paragraph{Consistency of Coplanar Repeats}
\label{sec:vp_estimation}
An affine-covariant region correspondence that is extracted from a
translational symmetry can provide point correspondences that are
translated in four directions in the scene \cite{Pritts-PAMI20}. See also \figref{fig:supp_vps_from_region} of the Supplemental. The
undistorted correspondences in each direction are used to estimate a
vanishing point incident to the vanishing line recovered by the minimal solver. This is computed by solving the
constrained least squares problem
\begin{equation}
    \underset{\vu}{\text{min}}~ \left\lVert\ma{M}\vu\right\rVert_2^2 \quad \text{subject
    to~} \ma{C}\vu = \ve[d],
\end{equation}
where \strut $\ma{C} = \begin{bmatrix} & \vl^{\T} & \\ 0 & 0 &
  1 \end{bmatrix}, \ve[d]= \colvec{2}{0}{1}$, $\ma{M}
  = \begin{bmatrix} \ve[t]_1~ \cdots
  ~\ve[t]_k \end{bmatrix}^\T$,\vspace{4pt} and
  $\ve[t]_1$,...,$\ve[t]_k$ are constructed by \eqref{eq:ct_meet}.
The point correspondences are used to
construct circles in the same way as contours, and the consistencies are measured by \eqref{eq:arc_vp_consistency}.  Point correspondences used to
compute the minimal solution will have zero error, but the minimal
solution can be cross-validated by points along the unused translation
directions provided by the region correspondence.

\section{Experiments}
\label{sec:evaluation}
The proposed solvers are quantitatively evaluated against the
state-of-the-art solvers listed in \tabref{tab:state_of_the_art} on
synthetic and real data.
Synthetic Manhattan scenes are used to assess the stabilities and
noise sensitivities of the solvers. The standard AIT dataset of
barrel-distorted images introduced by Wildenauer \etal in
\cite{Wildenauer-BMVC13} is used to assess the solver accuracy for the
auto-calibration task on challenging real images.

\begin{figure*}[!t]
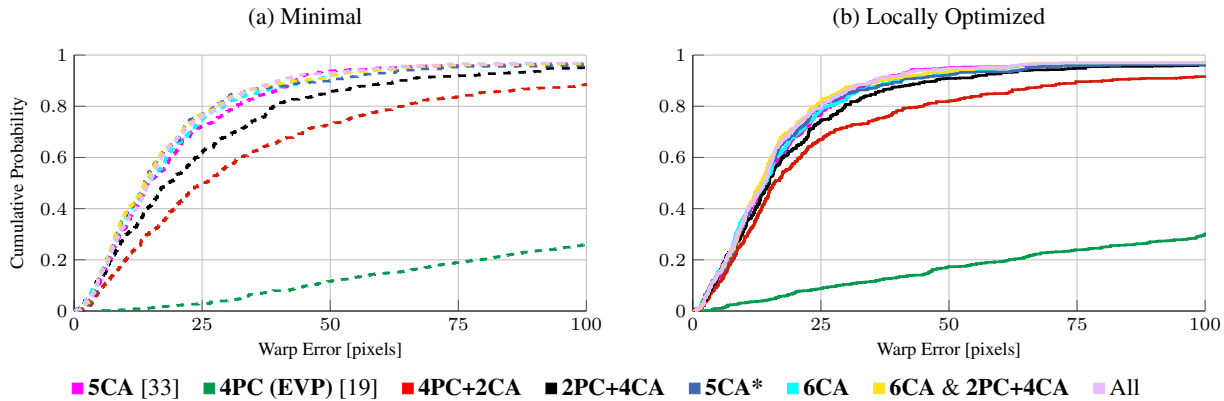

\centering
\setlength\fwidth{0.39\textwidth}
\subfloat[Minimal]{
\vspace{-7pt}
\scriptsize{\input{fig/AIT__min__warperr.tikz}}
\label{fig:AIT_minimal}}\hspace{10pt}
\subfloat[Locally Optimized]{
\vspace{-7pt}
\scriptsize{\input{fig/AIT__warperr.tikz}}
\label{fig:AIT_final}}\\
\small{\begin{tikzpicture}
\begin{customlegend}
[legend columns=-1,
legend style=
{draw=none,/tikz/every even column/.append style={column sep=0.25cm},cells={align=center}},
legend cell align={left},
legend entries={
\WildBMVC~\cite{Wildenauer-BMVC13},
\EVP~\cite{Pritts-CVPR18},
\HybridSolver,
\HybridSolverTwo,
\CircleDegSolver,
\CircleSolver,
\CircHyb,
\AllProposed
}
]
\addlegendimage{WildBMVC,fill=WildBMVC,only marks,mark=square*}
\addlegendimage{EVP,fill=EVP,only marks,mark=square*}
\addlegendimage{HybridSolver,fill=HybridSolver,only marks,mark=square*}
\addlegendimage{HybridSolverTwo,fill=HybridSolverTwo,only marks,mark=square*}
\addlegendimage{CircleDegSolver,fill=CircleDegSolver,only marks,mark=square*}
\addlegendimage{CircleSolver,fill=CircleSolver,only marks,mark=square*}
\addlegendimage{CircHyb,fill=CircHyb,only marks,mark=square*}
\addlegendimage{AllProposed,fill=AllProposed,only marks,mark=square*}
\end{customlegend}  
\end{tikzpicture}}
\vspace{-5pt}
\caption{Cumulative distributions of the warp error on AIT dataset~\cite{Wildenauer-BMVC13}. Results are shown for (a) the minimal \ie initial solutions and (b) locally optimized \ie final solutions. See also the distributions of the relative error of the division model parameter $\lambda$ and the focal length $f$ in \figref{fig:supp_AIT_stats} of Supplemental.}
\label{fig:AIT_stats}
\vspace{-15pt}
\end{figure*}

\subsection{Warp Error}
\label{sec:warp_error}
The warp error introduced in \cite{Pritts-BMVC16} is
adapted to jointly assess the accuracy of the estimated
 calibrations. Ground truth absolute orientation of the camera with
respect to the Manhattan frame is usually unknown for real images, thus
the performance is assessed based on the intrinsics. The image is
tessellated by an $N\times N$ grid of points
$\buildset{\vxd[i]}{i=1}{N^2}$. The tessellation ensures that the
error is uniformly sampled. The image points $\vxd[i]$ are
back-projected to rays using ground-truth. The rays are projected by
the estimated intrinsics and the reprojection error between the
projected rays and tessellation is used to define the warp error
\begin{equation} 
\label{eq:warp_error2}
\Delta_i = d(\vxd[i],g^{d}(\hat\mK \mK^{-1}g(\vxd[i],\lambda),\hat
\lambda)),
\end{equation}
where $d(\cdot,\cdot)$ is the Euclidean distance and $g^{d}$ is the
distortion transformation. The root mean square warp error for
$\buildset{\vxd[i]}{i=1}{N^2}$ is reported and denoted as
$\rmswarperr$. The warp error provides a geometric measure of
calibration accuracy; however, an error in focal length can be
compensated by an error in undistortion and vice-versa. Example warp
errors are illustrated in \secref{sec:supp_warp_error} of the
Supplemental.

\subsection{Numerical Stability}
\label{sec:stability}
The numerical stability measures the RMS warp error \rmswarperr of the
solvers on noiseless features. Configurations of coplanar mutually
orthogonal translated regions and parallel lines that are consistent
with each solver's required inputs are generated for realistic scenes
and camera configurations. \figref{fig:stability} reports the
distribution of $\log_{10}$ \rmswarperr on 1000 synthetic scenes. All
of the proposed solvers demonstrate good numerical stability, which is
consistent with the simple structure of the solvers. The arc-based
solver \WildBMVC of \cite{Wildenauer-BMVC13} has similar structure to
the proposed solvers and nearly as stable.  The \EVP solver of
\cite{Pritts-PAMI20} fails frequently. It is generated with the \Gbs
method, which solves a complicated system of polynomial equations.

\newcommand{\addrow}[1]{
\includegraphics[width=0.094\textwidth]{img/new/#1.JPG}
  &
  \includegraphics[width=0.143\textwidth]{img/new/EVP/#1_rect3.jpg}
  &
  \includegraphics[width=0.143\textwidth]{img/new/Wildenauer/#1_rect3.jpg}
  &
  \includegraphics[width=0.143\textwidth]{img/new/HybridSolver/#1_rect3.jpg}
  &
  \includegraphics[width=0.143\textwidth]{img/new/HybridSolverTwo/#1_rect3.jpg}
  &
  \includegraphics[width=0.143\textwidth]{img/new/CircleDegSolver/#1_rect3.jpg}
  &
  \includegraphics[width=0.143\textwidth]{img/new/CircleSolver/#1_rect3.jpg}
  \\
}
\renewcommand{\tabcolsep}{2pt}
\newcolumntype{P}[1]{>{\centering\arraybackslash}m{#1}}
\begin{figure*}[!t]
\centering
\captionsetup[subfigure]{labelformat=empty}
{\def\arraystretch{0.7}
\begin{tabularx}{\textwidth}{@{}P{1.54cm} *6{P{2.47cm}}}
\small Input &
\small \EVP \cite{Pritts-CVPR18} &
\small \WildBMVC \cite{Wildenauer-BMVC13} &
\small \HybridSolver &
\small \HybridSolverTwo &
\small \CircleDegSolver &
\small \CircleSolver\\[5pt]
\addrow{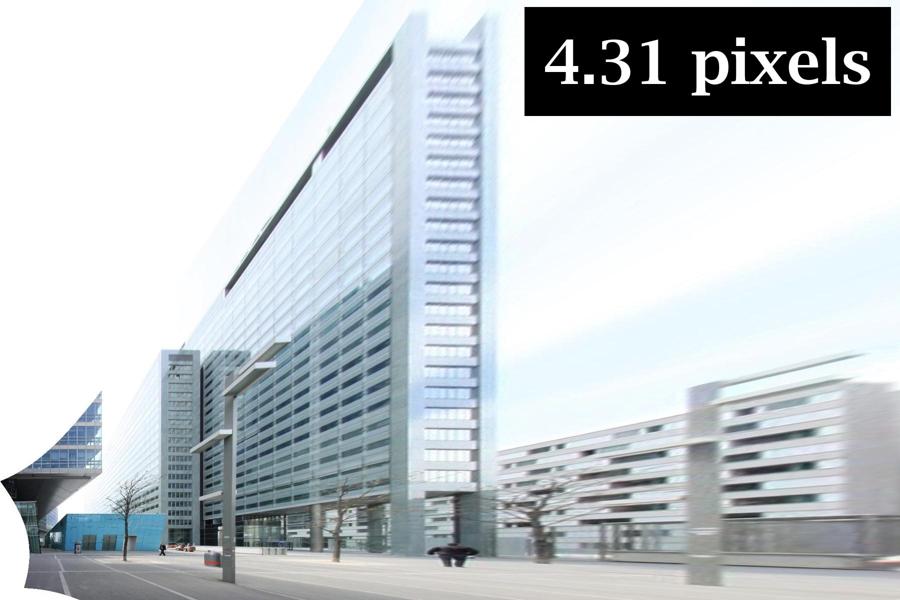}\\[-10pt]
\addrow{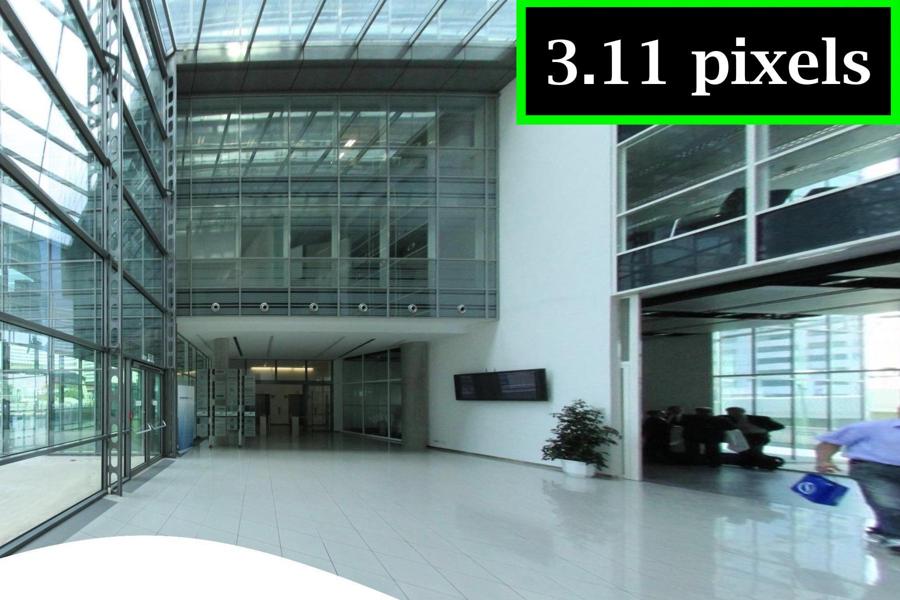}\\[-10pt]
\addrow{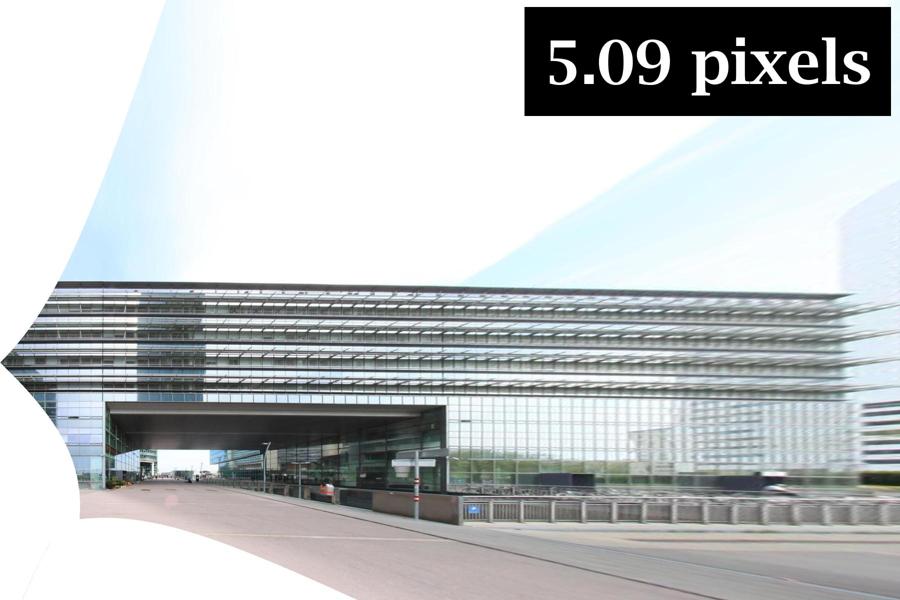}\\[-10pt]
\addrow{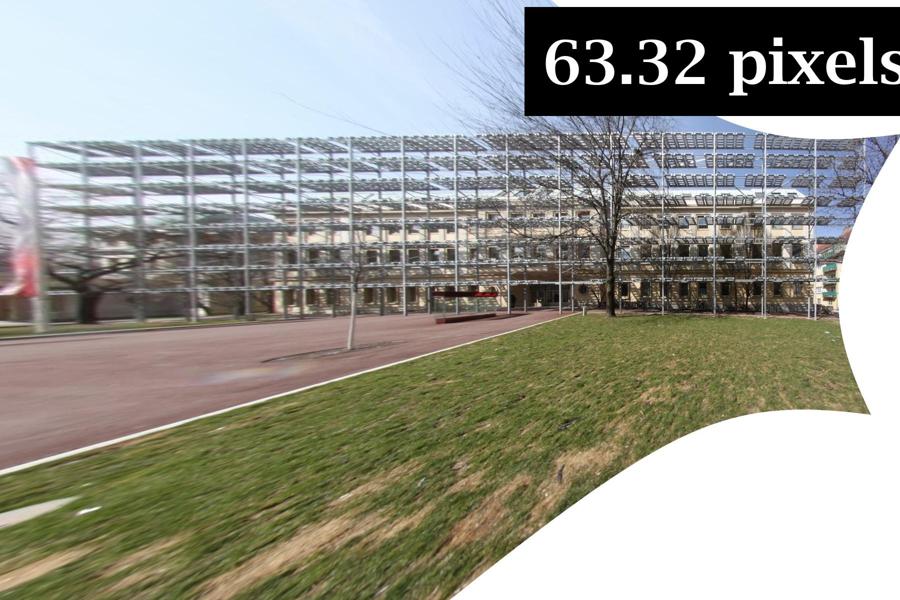}\\[-10pt]
\end{tabularx}\vspace{-5pt}
}
\caption{Example scenes for which either the
  arc-based or hybrid solvers have an advantage. The input image is on
  the left, and the dominant plane metrically-rectified for each
  solver is on the right. RMS warp errors are reported in the top right corners. Best results are
  in green.}
\label{fig:real_data_AIT}
\vspace{-15pt}
\end{figure*}

\subsection{Noise Sensitivity}
\label{sec:sensitivity}
The proposed and state-of-the-art solvers are evaluated for their
robustness to sensor noise (see \figsref{fig:sensitivity_circles} and
\ref{fig:sensitivity_rgns}).  White noise generated from
$\mathcal{N}(\ve[0]_2, \sigma^2 \ma{I}_2)$ is added to the imaged
translated symmetries and parallel lines. Solver sensitivity is
measured at noise levels of $\sigma \in \buildset{ 0.1, 0.5,1,2
}{}{}$.  The solvers are used in a basic \RANSAC estimator that
minimizes the RMS warp error \rmswarperr over 25 minimal samples for
1000 scenes at each noise level. As expected, the solvers admitting
arcs give superior performance since the entire contour is used to
regress the arc. The proposed \CircleSolver is the most robust, while
the proposed \CircleDegSolver and state-of-the-art \WildBMVC solver
\cite{Wildenauer-BMVC13} are both competitive. The proposed
\HybridSolver is significantly more robust than the state-of-the-art
\EVP solver \cite{Pritts-CVPR18}, and the proposed \HybridSolverTwo is
competitive with the arc-based solvers.

\subsection{Real Data}
\label{sec:real_data}
The solvers are evaluated on the AIT dataset introduced in
\cite{Wildenauer-BMVC13}, which consists of 102 barrel-distorted
images taken by Canon EOS 500D camera with a Walimex Pro 8mm
$170^{\circ}$ HFOV fisheye lens. An offline calibration is provided
with the dataset and used as the ground truth. The image center,
distortion center, and principal point are assumed to be the same, and
lens undistortion is modeled with the division model. The calibrated
focal length is $1126.3$ pixels and division model parameter is
$-2.4951\times10^{-7}$ pixels for the AIT images at a resolution of
$3000\times2000$.

The number of iterations was set to $1000$ for the region based \EVP solver, and $500$ for the other solvers. The reprojection threshold is $5.05$ pixels for points of covariant regions and $1.26$ pixels for contour points.
The consistency of the features with
an auto-calibration used for evaluation is similar to
\eqref{eq:arc_vp_consistency}, and the models with maximal consensus
sets are locally optimized by a method similar to
\cite{Pritts-CVPR14}.

Auto-calibrations on the AIT dataset are accumulated over ten
runs. \tabref{tab:AIT} reports the percentage of Top-1 solutions
achieved by each solver, where a Top-1 solution is the
auto-calibration that has the minimal RMS warp error for an image. We
also report the median relative focal length and lens undistortion
errors and the median RMS warp error.  The best performing arc-based
solver is the proposed \CircleSolver with twice as many Top-1
solutions and a $2\%$ reduction in the warp error compared to the
state-of-the-art \WildBMVC solver. The \HybridSolverTwo hybrid solver
is the best performing for solvers using points, and it significantly
outperforms the state-of-the-art \EVP.
As shown by \tabref{tab:AIT}, subsets of the AIT dataset are best
solved by particular solvers, which suggests that using combinations
of solvers is necessary to recover the best calibration across the
dataset. Hybrid RANSAC was used to sample arcs and points according
to the input types listed in \tabref{tab:state_of_the_art}
\cite{Camposeco-CVPR18}. The solver admitting the sampled input is
invoked to hypothesize an auto-calibration. We used the following
solvers together: \begin{enumerate*}[(i)] \item all hybrid
solvers, \item all arc-based solvers \item all solvers, and \item
only the \CircleSolver and \HybridSolverTwo solvers.\end{enumerate*}
The configuration using all solvers gave the most accurate estimates
of focal length and division model parameter, while the combination of
\CircleSolver and \HybridSolverTwo gave the lowest warp error.

\figref{fig:AIT_stats} reports the cumulative distributions of errors
reported in \tabref{tab:AIT} for the individual solvers and for the
combination of all solvers, as well as using the \CircleSolver and
\HybridSolverTwo in combination. The proposed \CircleDegSolver,
\CircleSolver, \HybridSolverTwo and the state-of-the-art \WildBMVC of
\cite{Wildenauer-BMVC13} give comparably good performance. The
region-based \EVP solver of \cite{Pritts-CVPR18} performs poorly, due
to noisy covariant region detections. The solver combinations are better than all individual results. Notably, the
performance of the \CircleSolver and \HybridSolverTwo combination
matches the use of all solvers and exceeds the performance of the
individual solvers. This suggests that the dataset is saturated if
only arc-based solvers are used, and the inclusion of a point-based
solver is necessary to improve calibration accuracy. Inclusion of more
solvers gives only a small increase in accuracy across the dataset,
which may show up in Top-1 solutions, but is not impactful enough to
justify adding extra solvers. Choosing the two best individual arc and
hybrid solvers---\CircleSolver and \HybridSolverTwo---works well.

\figref{fig:real_data_AIT} shows diverse scene content for which
either the arc-based or hybrid solvers have an advantage.
Note that the scenes where the \HybridSolver and \HybridSolverTwo
hybrid solvers perform best have high-contrast lines in only one
direction, which suggests that translational symmetries are needed to
constrain the second Manhattan frame direction.

\section{Conclusions}
\label{sec:conclusions}
We propose rectifying and calibrating minimal solvers that use
combinations of circular arcs and point correspondences, which are
images of parallel lines and translational symmetries in the scene,
respectively. The proposed solvers extend accurate rectification and
auto-calibration to distorted images of scenes that lack either
coplanar texture or parallel scene lines. No individual solver emerged
as a clear winner for the task of auto-calibration on the standard AIT
dataset. Instead, we found that the solvers are complementary. Each
solver works best on scenes with content that reflect its particular
input configuration. There is a benefit to including the constraints
from imaged translational symmetries even though the point
correspondences extracted from covariant regions are 
more noisy than the circular arcs. However, experiments suggest that a
significant improvement could be achieved by refining translational
symmetries detected from covariant region correspondences. We achieve
state-of-the-art performance on the AIT dataset by using multiple
solvers in a \RANSAC variant that samples different combinations of
point correspondences and arcs as input for each of the solvers.

{\small
\bibliographystyle{ieee_fullname}
\bibliography{wacv21}
}

\vspace{5pt}
\begingroup
\let\cleardoublepage\relax
\let\clearpage\relax
\onecolumn
\begin{center}
\textbf{
\Large Minimal Solvers for Single-View Lens-Distorted Camera Auto-Calibration\\[14pt]
Supplemental Material}
\end{center}
\twocolumn
\endgroup
\normalsize
\setcounter{section}{0}
\counterwithin{figure}{section}
\counterwithin{table}{section}
\renewcommand\thesection{\Alph{section}}

\vspace{16pt}
\section{Geometry of Input Features}
\label{sec:supp_bmss}

As discussed in \secref{sec:consistency_measure}, the input configurations of the solvers provide
extra measurements that can be used to reject invalid solutions. Circular arcs provide contour points that can be tested against the vanishing point. Region correspondences provide point correspondences in three other translational directions that can be tested against the vanishing lines. In the following sections, we discuss the construction of the circles and vanishing points.

\subsection{Circle through Vanishing Point}
\label{sec:supp_circle_through_vp}
\begin{figure}[h!]
\centering
\includegraphics[width=0.75\columnwidth]{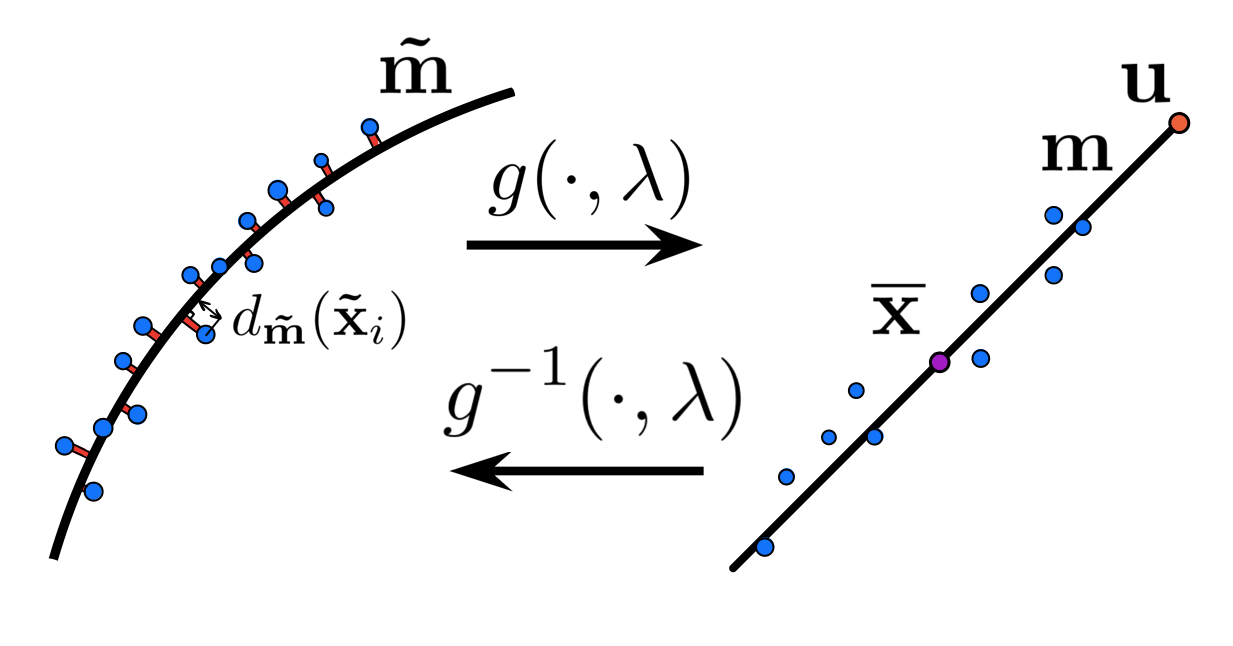}
\mycaption{Geometry of the Consistency Measure}{Left is the distorted space, and right is the undistorted space. Midpoint $\overline\vx$ and line $\ve[m]$ is constructed in undistorted space and then warped to distorted space. Mean-squared distance from the points to $\ve[\tilde m]$ is computed in the distorted space.}
\label{fig:supp_circle_through_vp}
\end{figure}

The construction of the consistency measure \eqref{eq:arc_vp_consistency} is a generalization of the Tardif consistency measure for
vanishing points and imaged parallel lines introduced
in \cite{Tardif-ICCV09}. Vanishing point $\ve[u]$ is recovered from a minimal solution. The join $\ve[m]$
of $\vu$ with the undistorted midpoint $\overline\vx$ of 
either the circular arc or the point correspondence is constructed (see the right part of \figref{fig:supp_circle_through_vp}). Line
$\ve[m]$ is distorted to a circle $\ve[\tilde m]$ using the minimal
solution of the division model parameter, and the mean squared
distances of image points to the circle $d_{\ve[\tilde m]}(\vxd[i])$
is computed (see the left part of \figref{fig:supp_circle_through_vp}).

\subsection{Vanishing Points from a Region Correspondence}
\label{sec:supp_vps_from_region}
\begin{figure}[h!]
\centering
\includegraphics[width=\columnwidth]{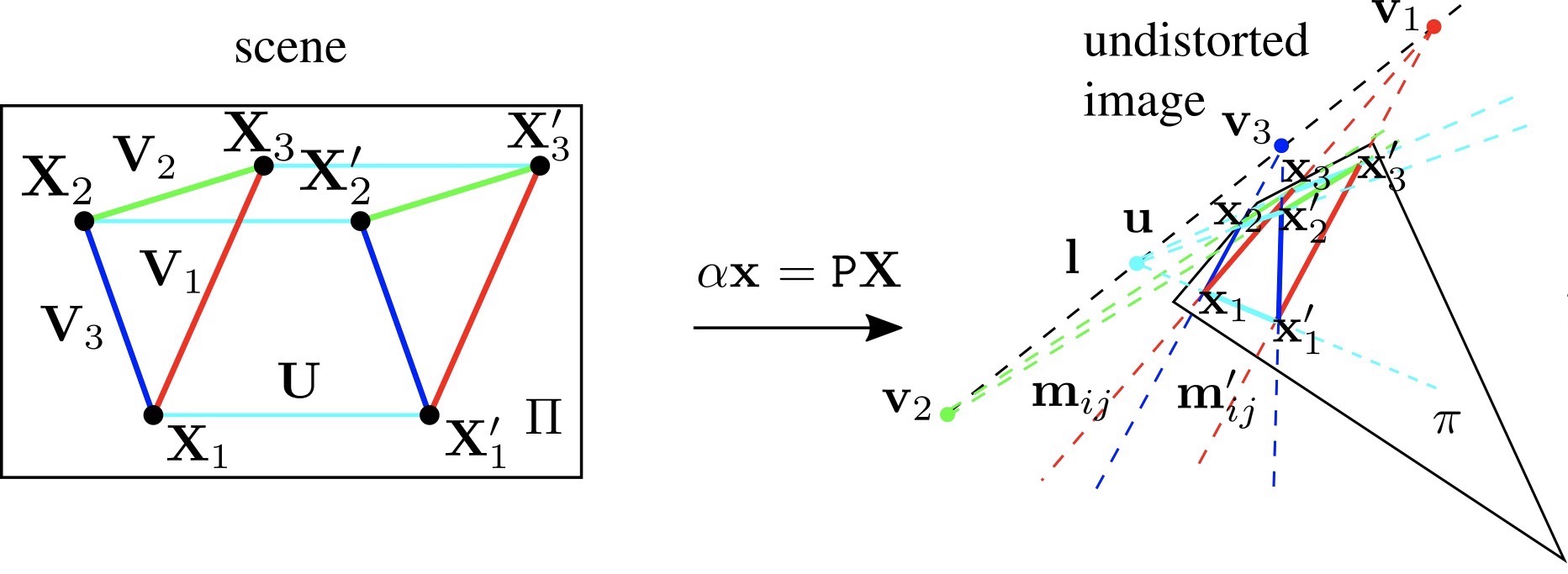}
\mycaption{Geometry of an Affine-Covariant Region}{The scene 
plane $\Pi$ contains the preimage of radially-distorted
conjugately-translated affine-covariant regions, equivalently, 3
translated points in the direction \vU. This configuration had 3
additional translation directions \vV[1],\vV[2],\vV[3] that can be
used to design a solver or to validate a minimal solution. Courtesy of \cite{Pritts-PAMI20}.}
\label{fig:supp_vps_from_region}
\end{figure}

\figref{fig:supp_vps_from_region} shows two
corresponded affine-covariant regions in two spaces: scene space and undistorted image space. In particular, affine-covariant
regions are parameterized by affine frames (defined by three points),
which is a common parameterization. Six point correspondences can be
extracted from the region correspondence.

\newpage
\quad
\vspace{52pt}

The orientations of the joins of points are color coded. Four vanishing points can be constructed from the red, green, blue and cyan lines. Up to two vanishing points are constructed by the solvers \HybridSolver and \HybridSolverTwo. The remaining vanishing points can be estimated by constraining them to lie on the recovered vanishing line $\vl$. Then the vanishing points can be used to validate the minimal solution of the division model parameter $\lambda$ and $\vl$ using the consistency measure \eqref{eq:arc_vp_consistency}.

\begin{figure}[!h]
\centering
\includegraphics[width=\columnwidth]{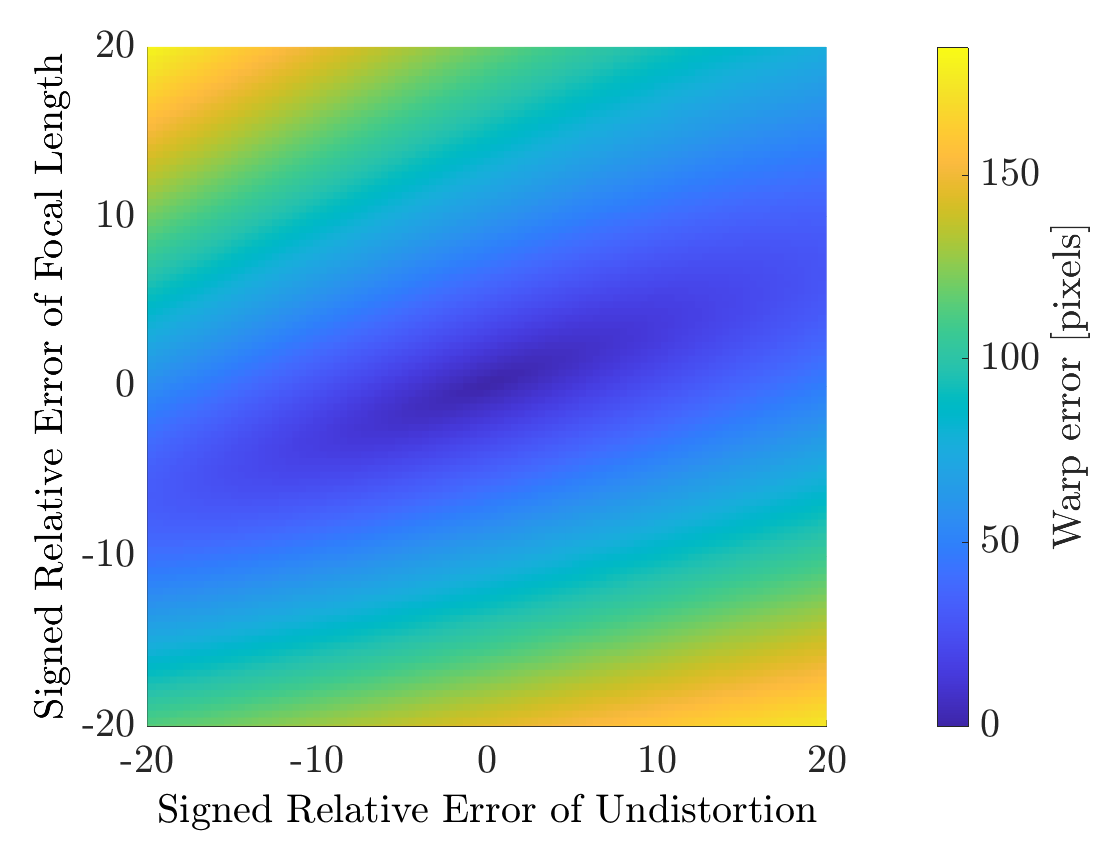}
\mycaption{Warp Error as a Function of the Relative Errors of Undistortion and Focal Length}{The warp error is calculated for $100\times100$ tessellation of $[-20,20]\times[-20,20]$ space of the relative errors of undistortion and focal length.}
\label{fig:warperr}
\end{figure}

\begin{table*}[!t]
\setlength\cellspacetoplimit{1.4pt}
\setlength\cellspacebottomlimit{1.4pt}
\newcolumntype{P}[1]{>{\centering\arraybackslash}m{#1}}
\centering
\begin{tabularx}{\textwidth}{@{}*2{c@{}}*5{S{Y}@{}}}
& & 
\multicolumn{5}{c}{Signed Relative Error of Undistortion}
\\
& &
$-10\%$ &
$-5\%$ &
$0$ &
$5\%$ &
$10\%$ \\
\multirow{5}{*}{\rotatebox{90}{Signed Relative Error of Focal Length\qquad\qquad\qquad}} &
$-10\%$ &
\includegraphics[height=2.34cm,valign=M]{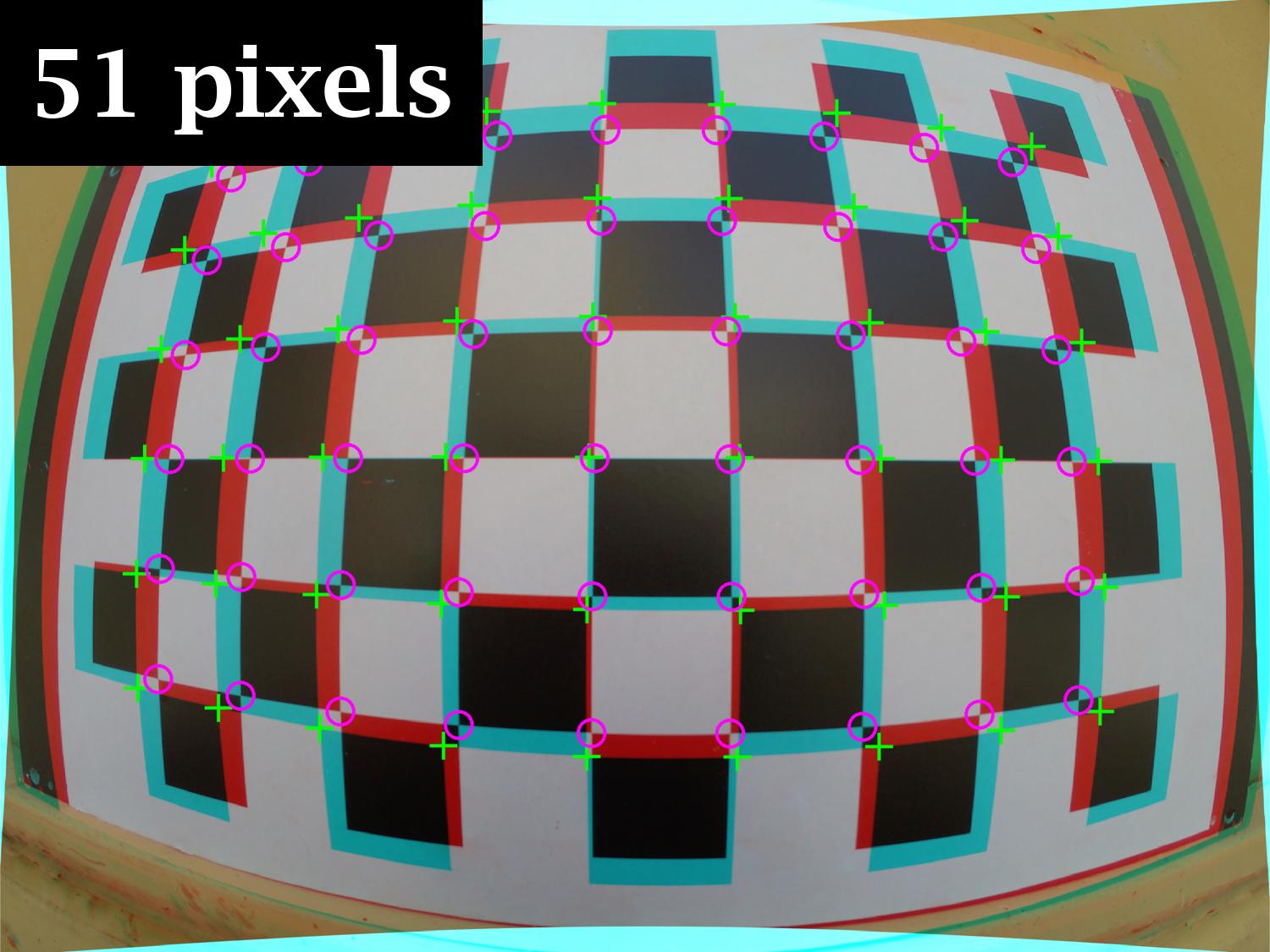} &
\includegraphics[height=2.34cm,valign=M]{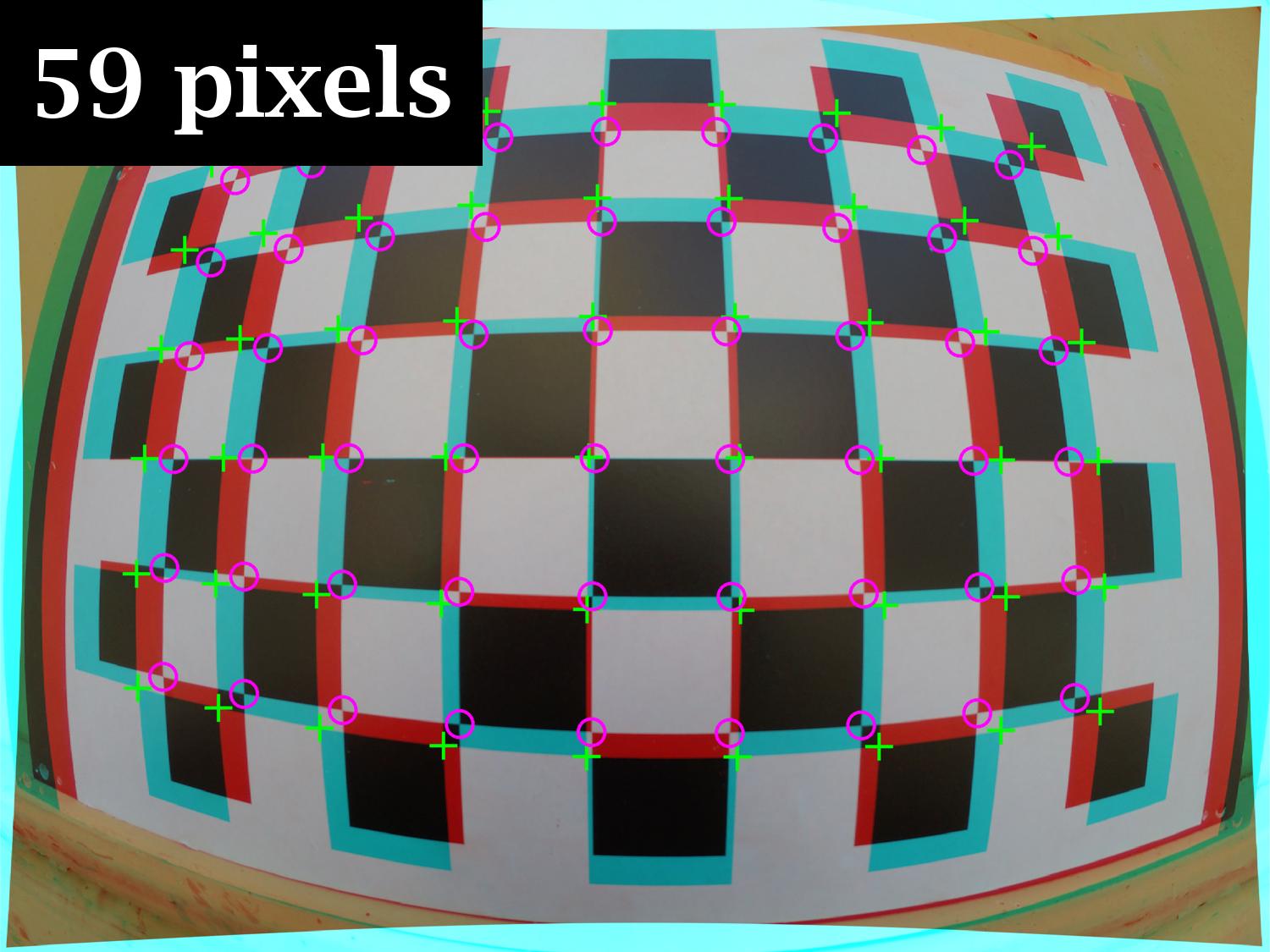} &
\includegraphics[height=2.34cm,valign=M]{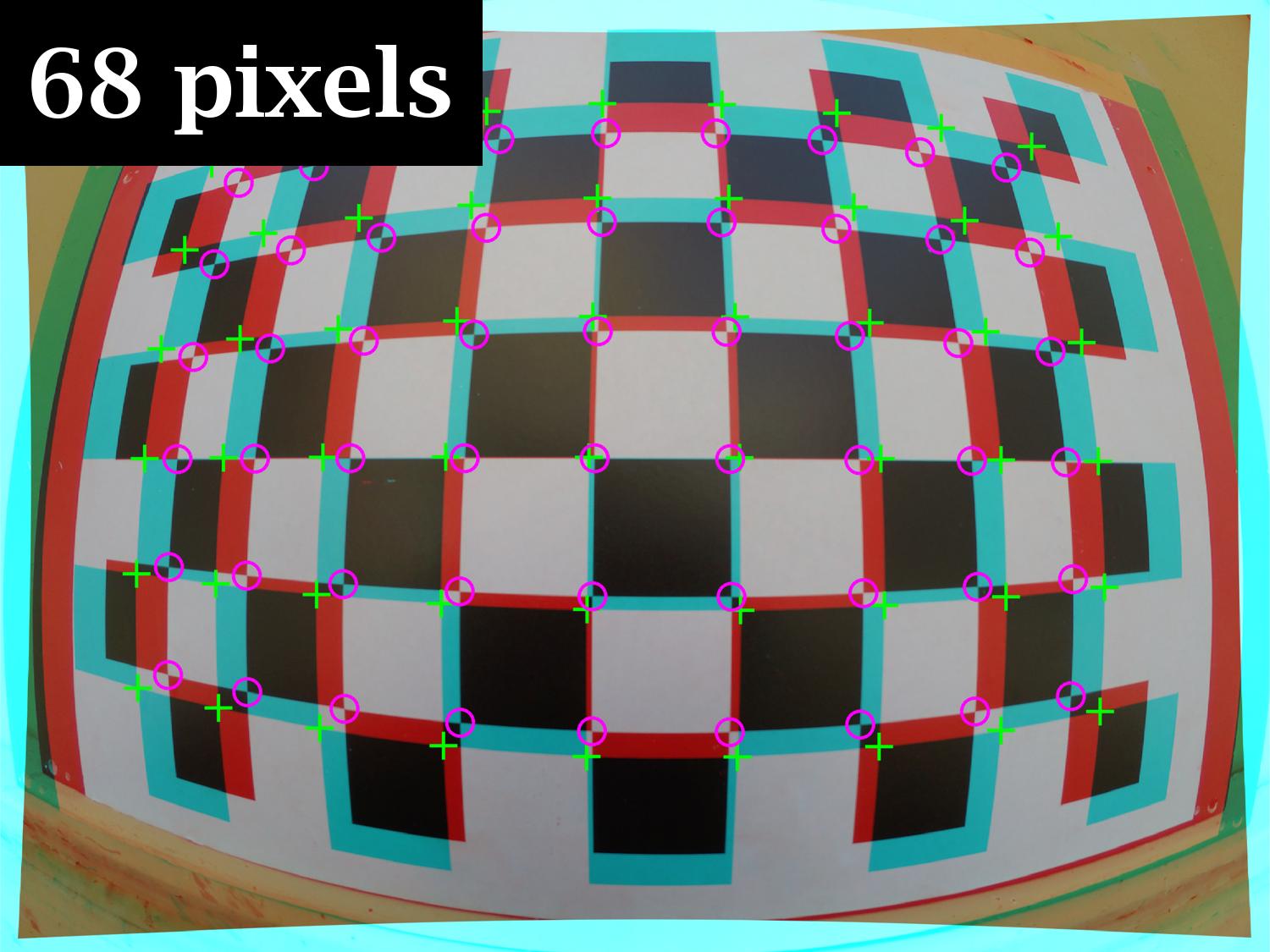} &
\includegraphics[height=2.34cm,valign=M]{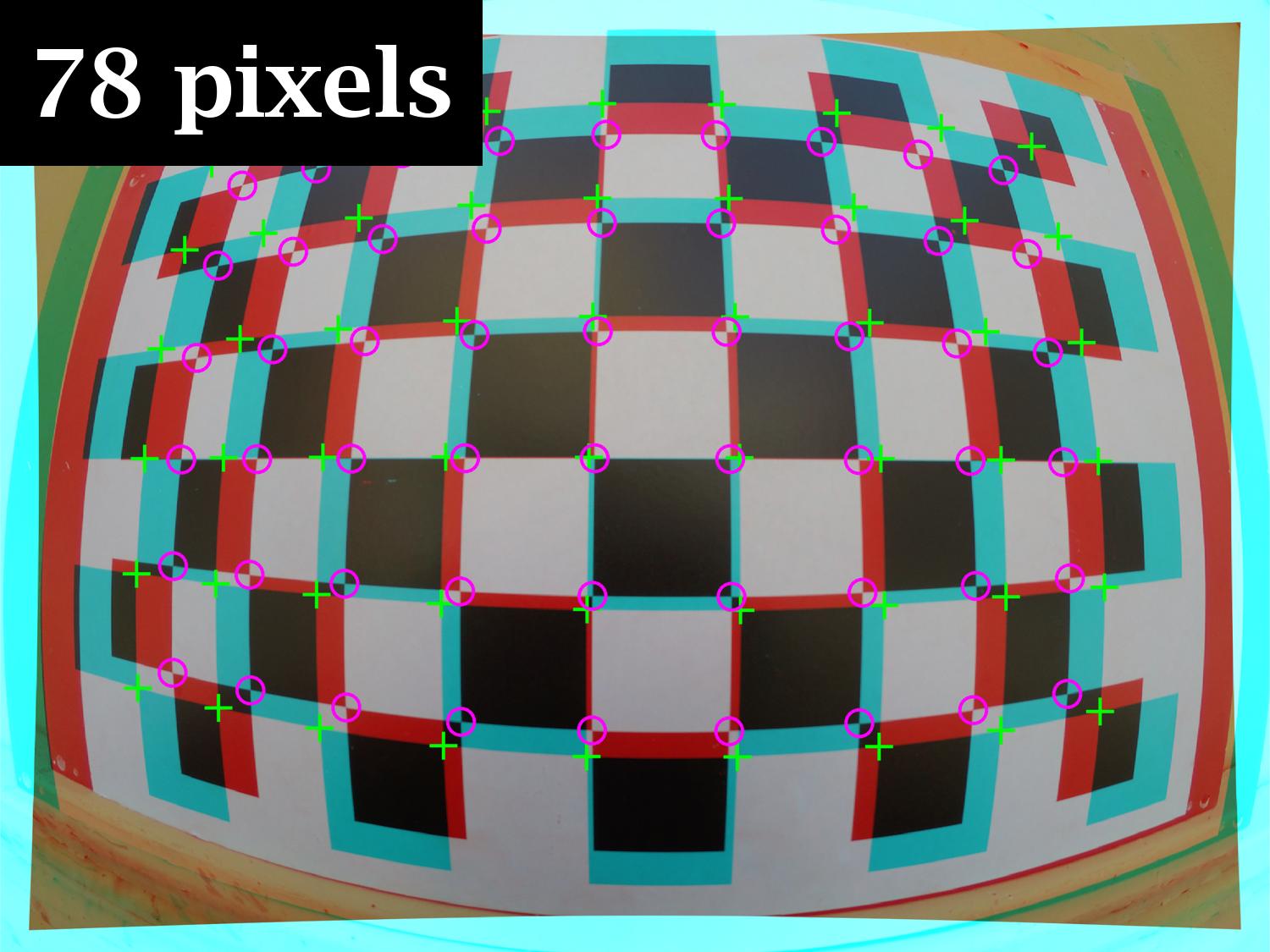} &
\includegraphics[height=2.34cm,valign=M]{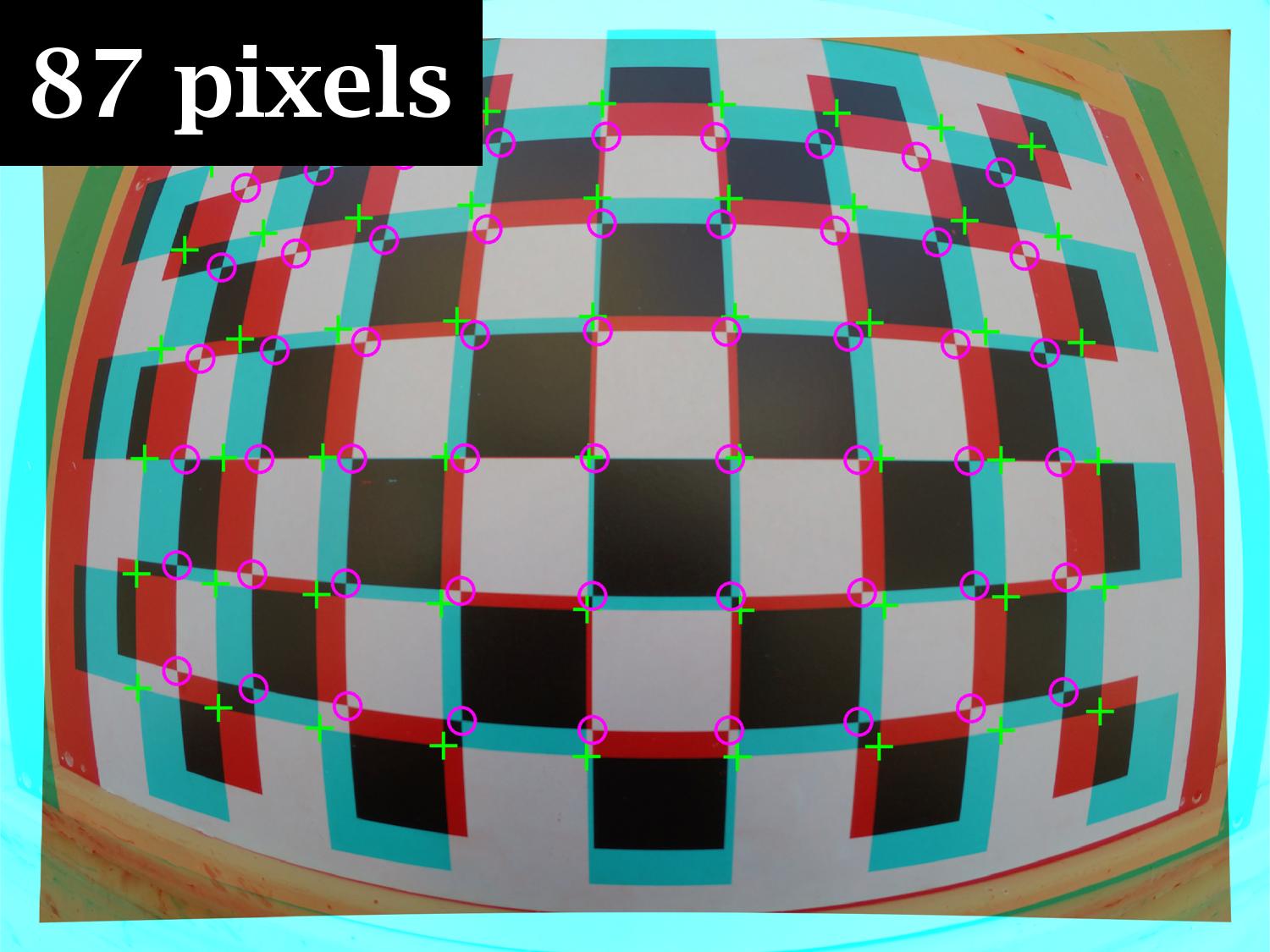} \\
& $-5\%$ &
\includegraphics[height=2.34cm,valign=M]{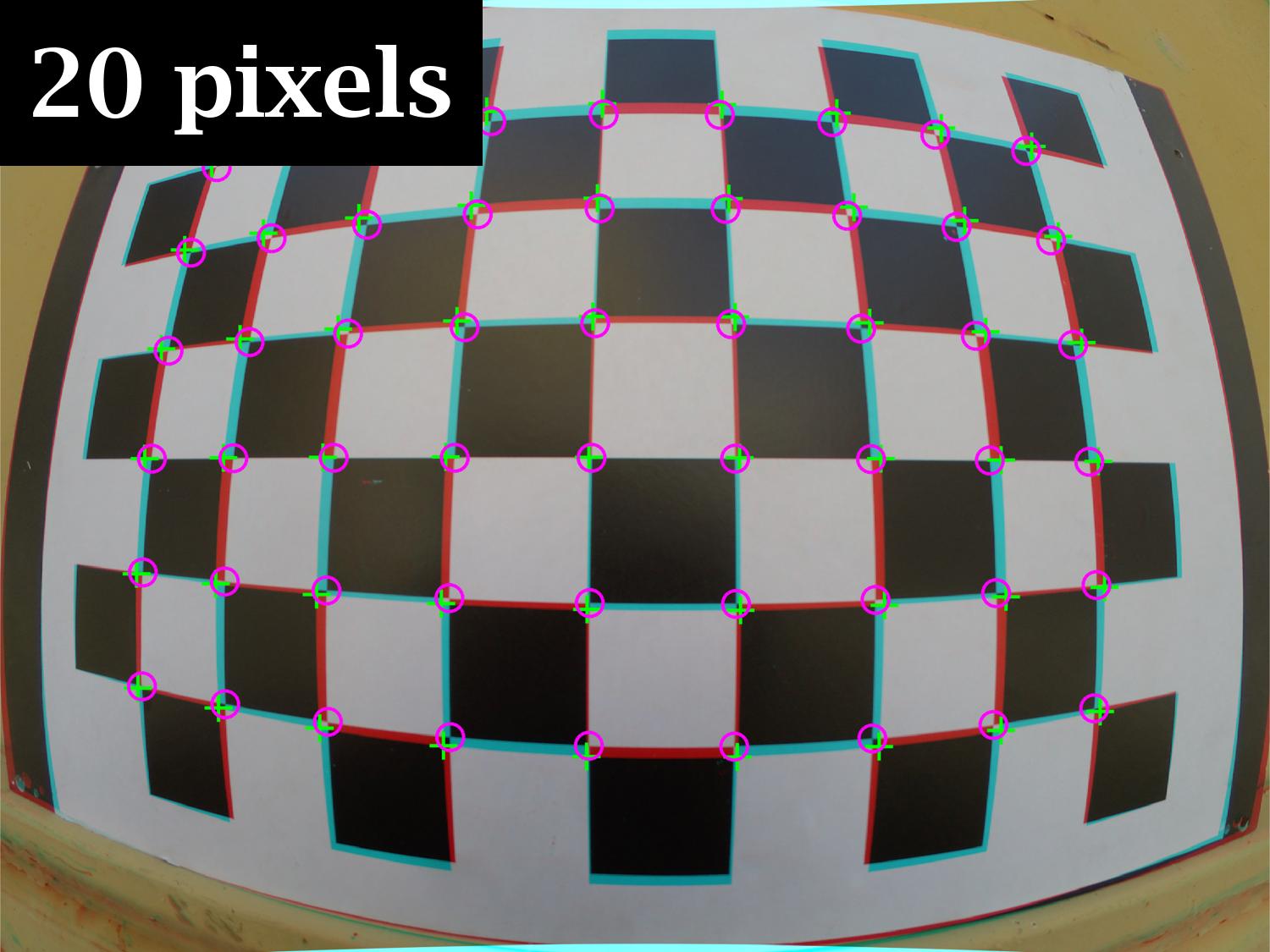} &
\includegraphics[height=2.34cm,valign=M]{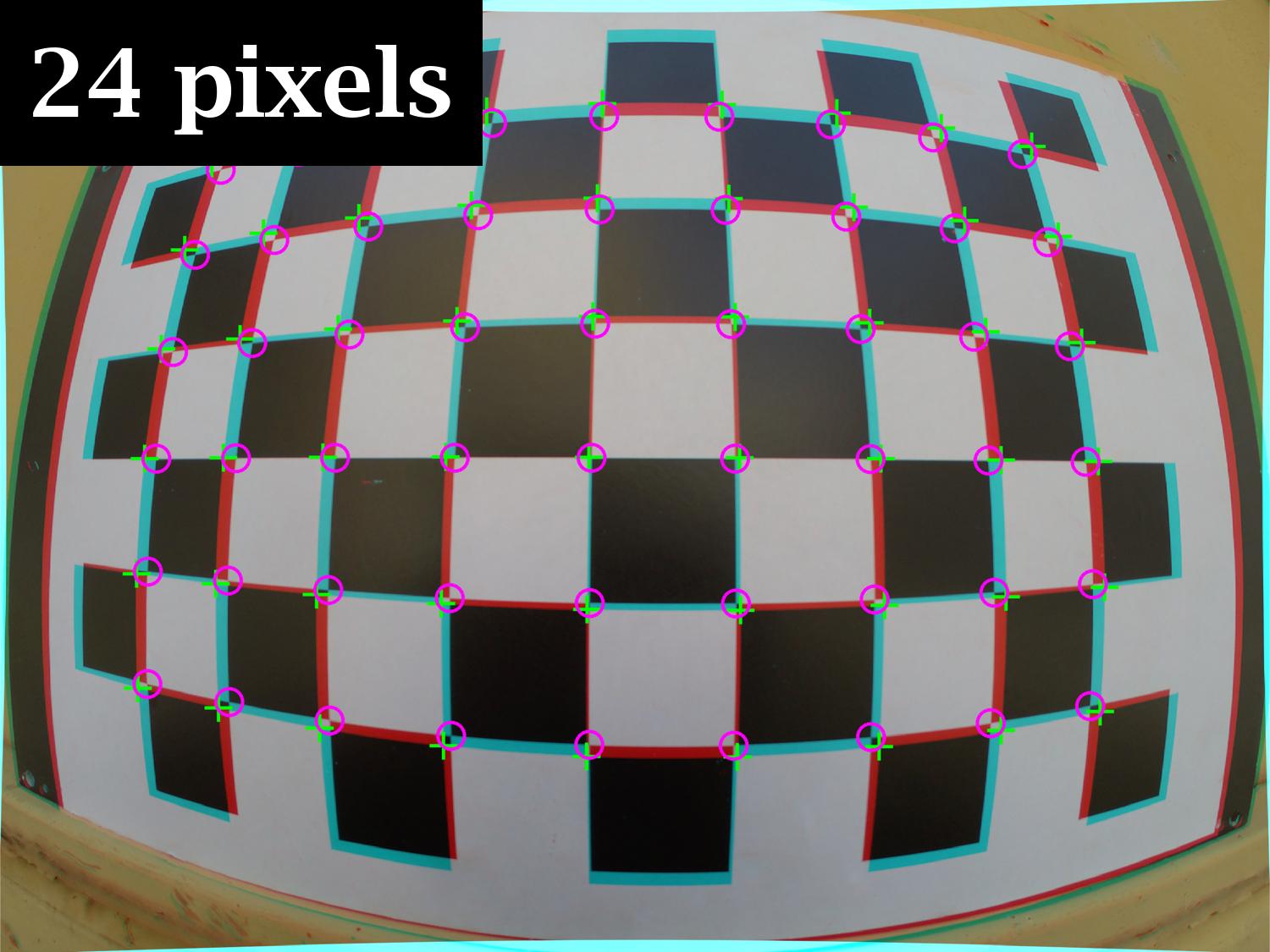} &
\includegraphics[height=2.34cm,valign=M]{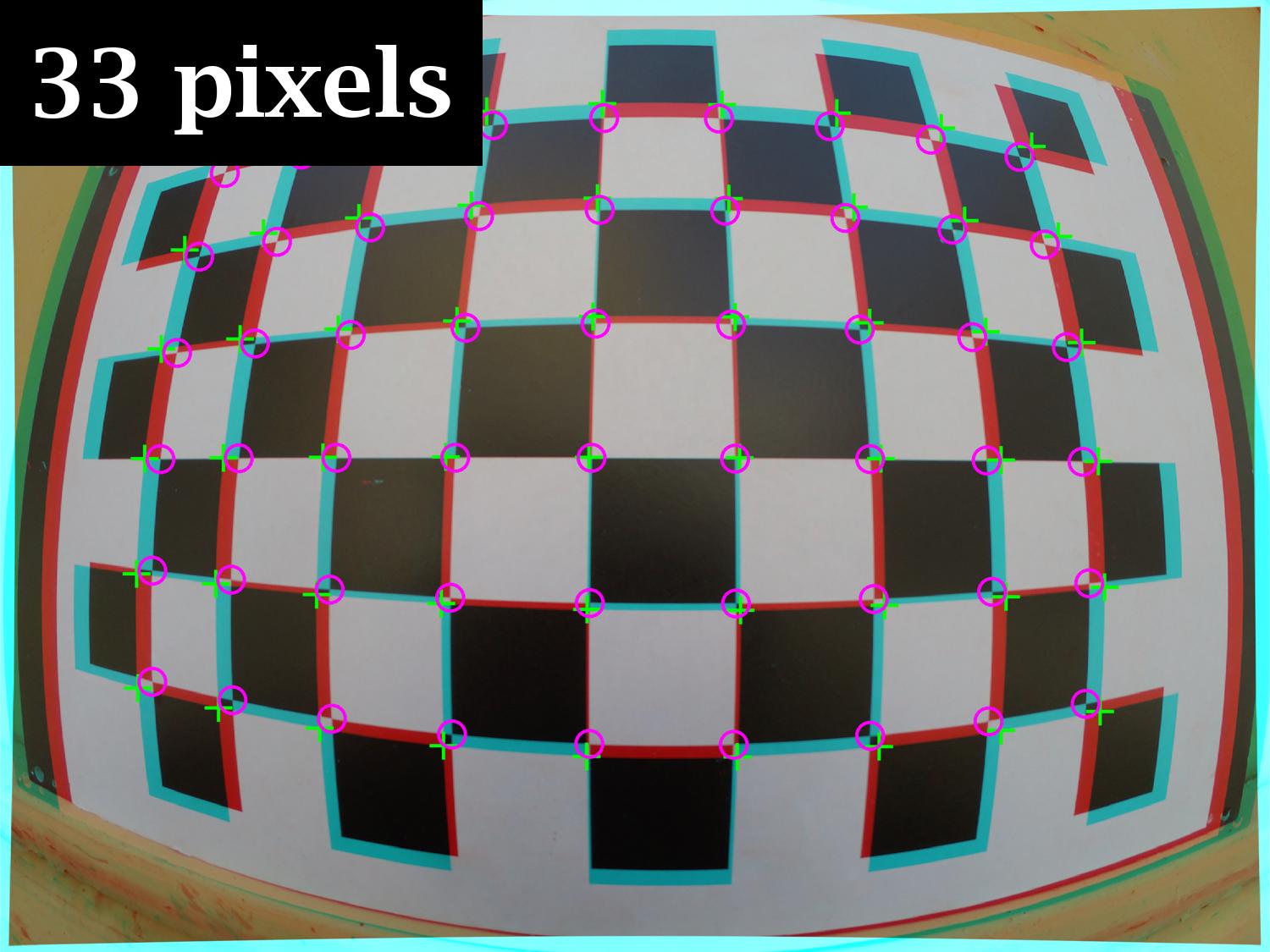} &
\includegraphics[height=2.34cm,valign=M]{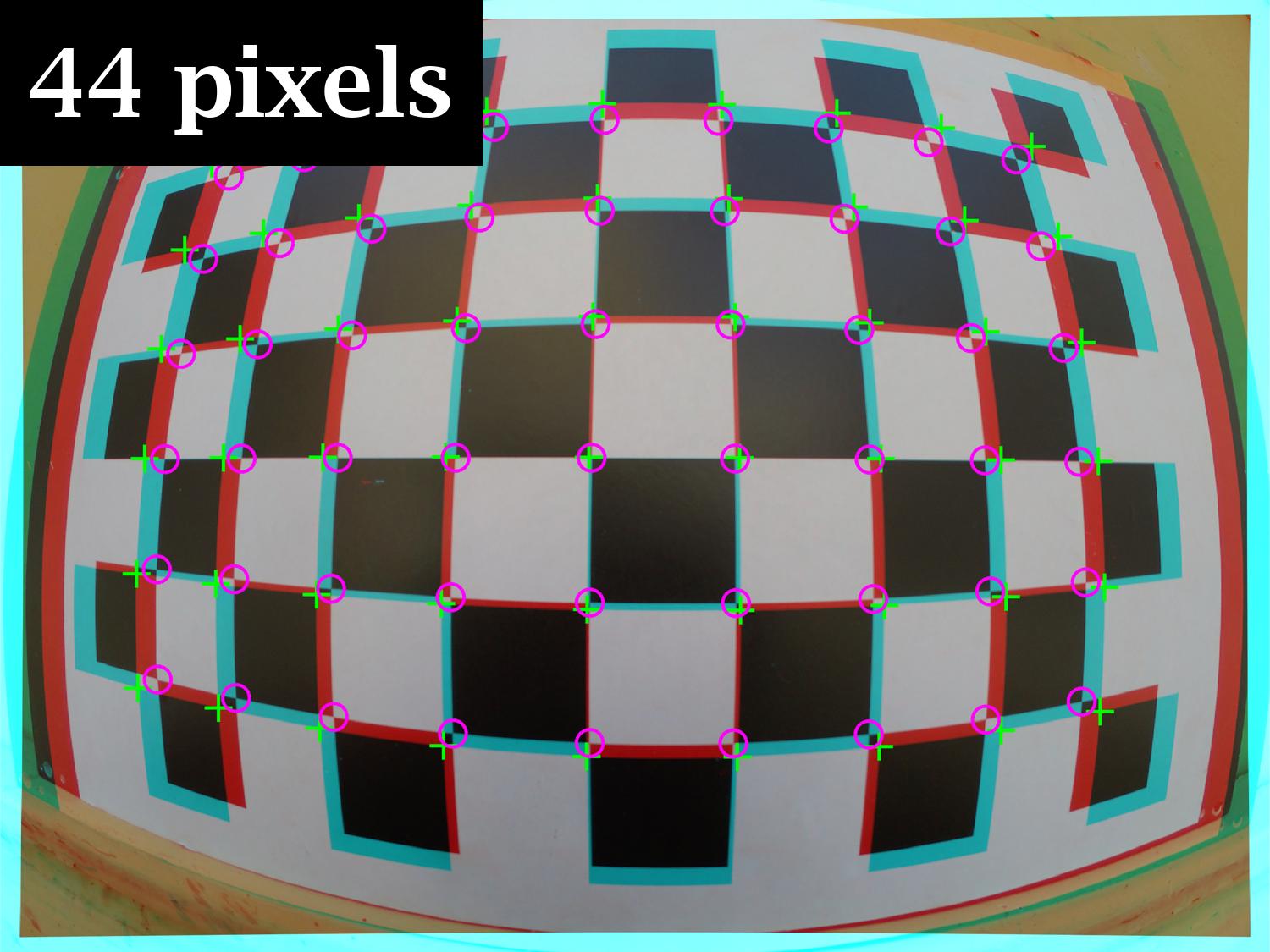} &
\includegraphics[height=2.34cm,valign=M]{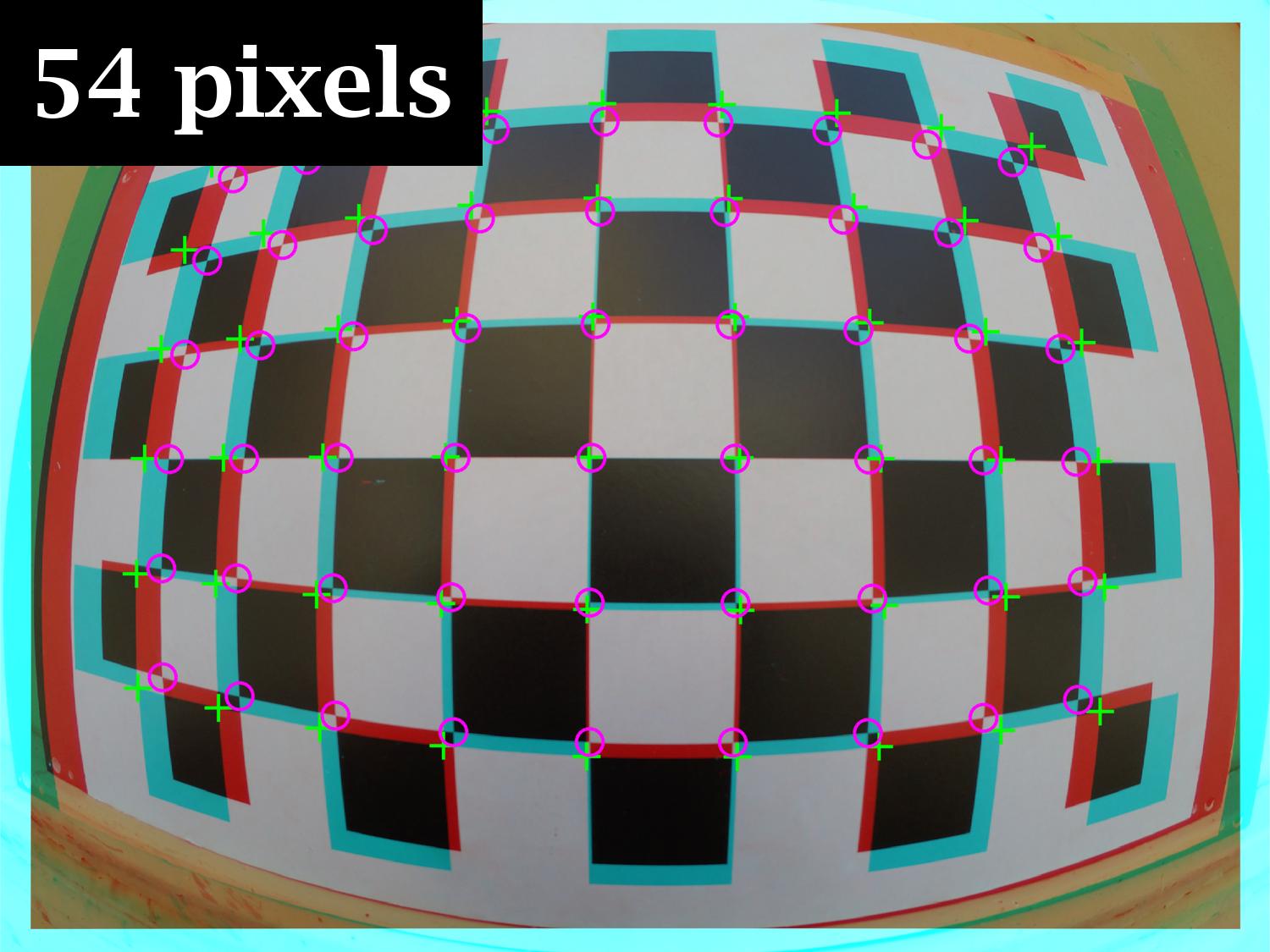} \\
& $0$ &
\includegraphics[height=2.34cm,valign=M]{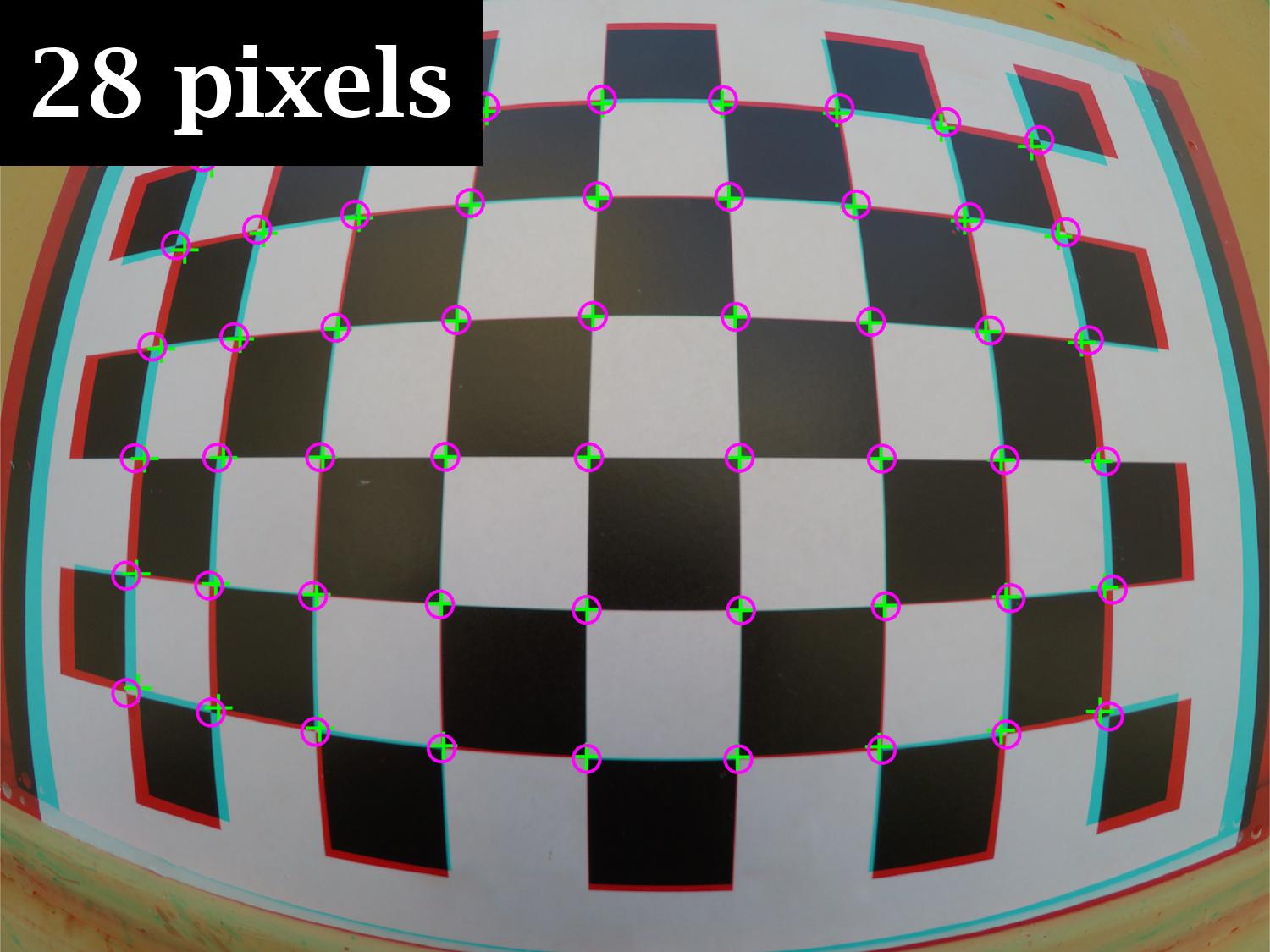} &
\includegraphics[height=2.34cm,valign=M]{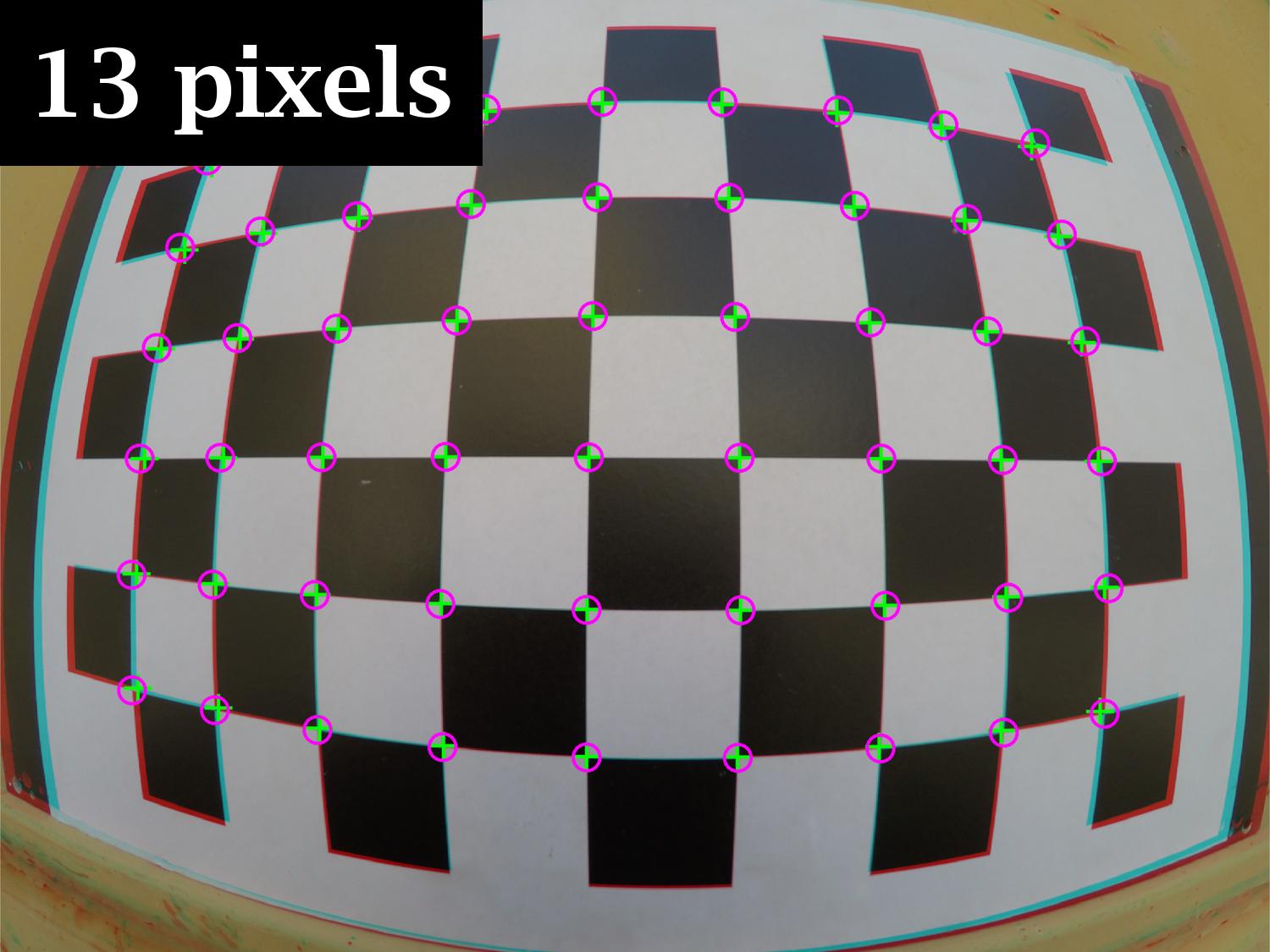} &
\includegraphics[height=2.34cm,valign=M]{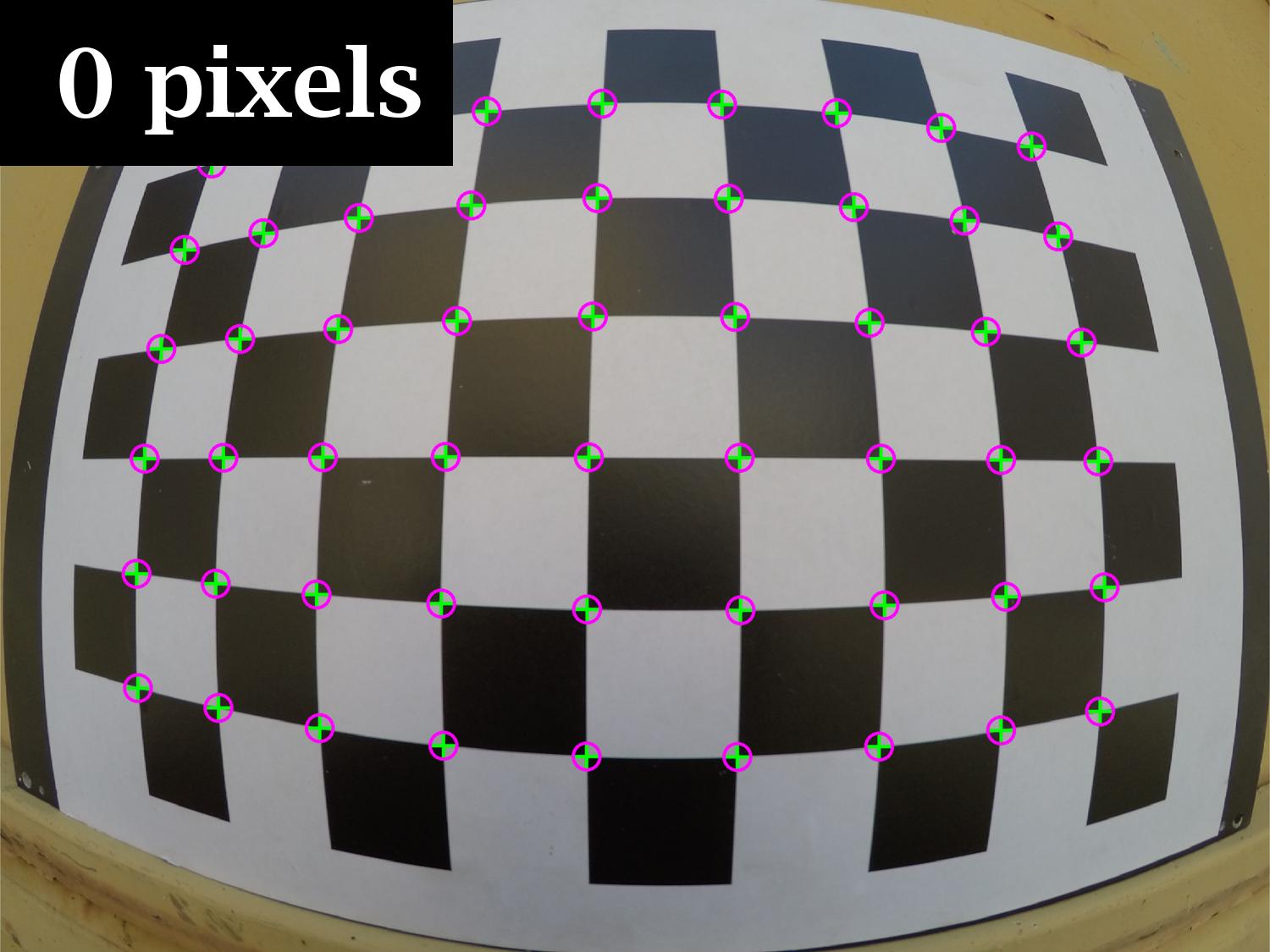} &
\includegraphics[height=2.34cm,valign=M]{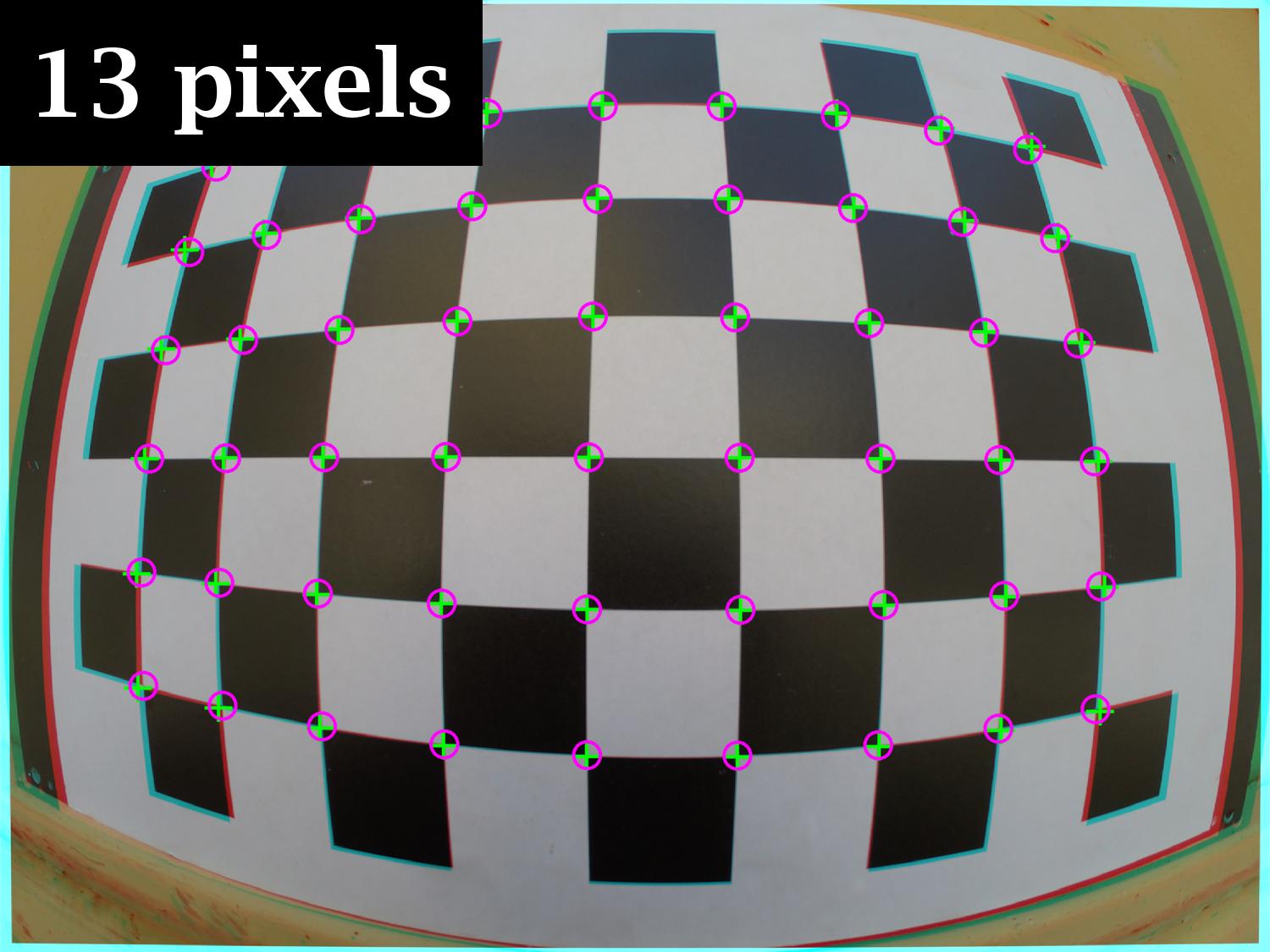} &
\includegraphics[height=2.34cm,valign=M]{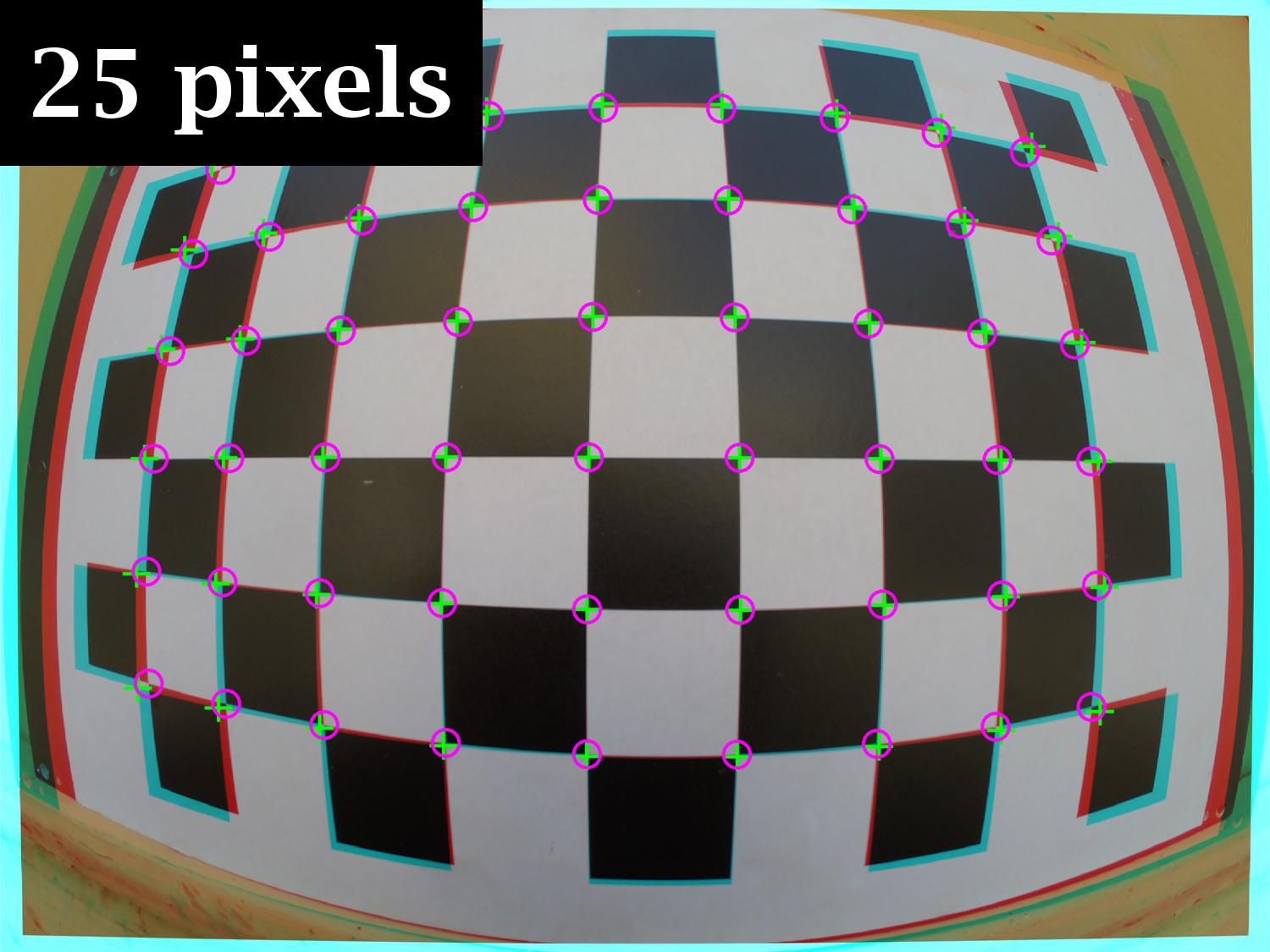} \\
& $5\%$ &
\includegraphics[height=2.34cm,valign=M]{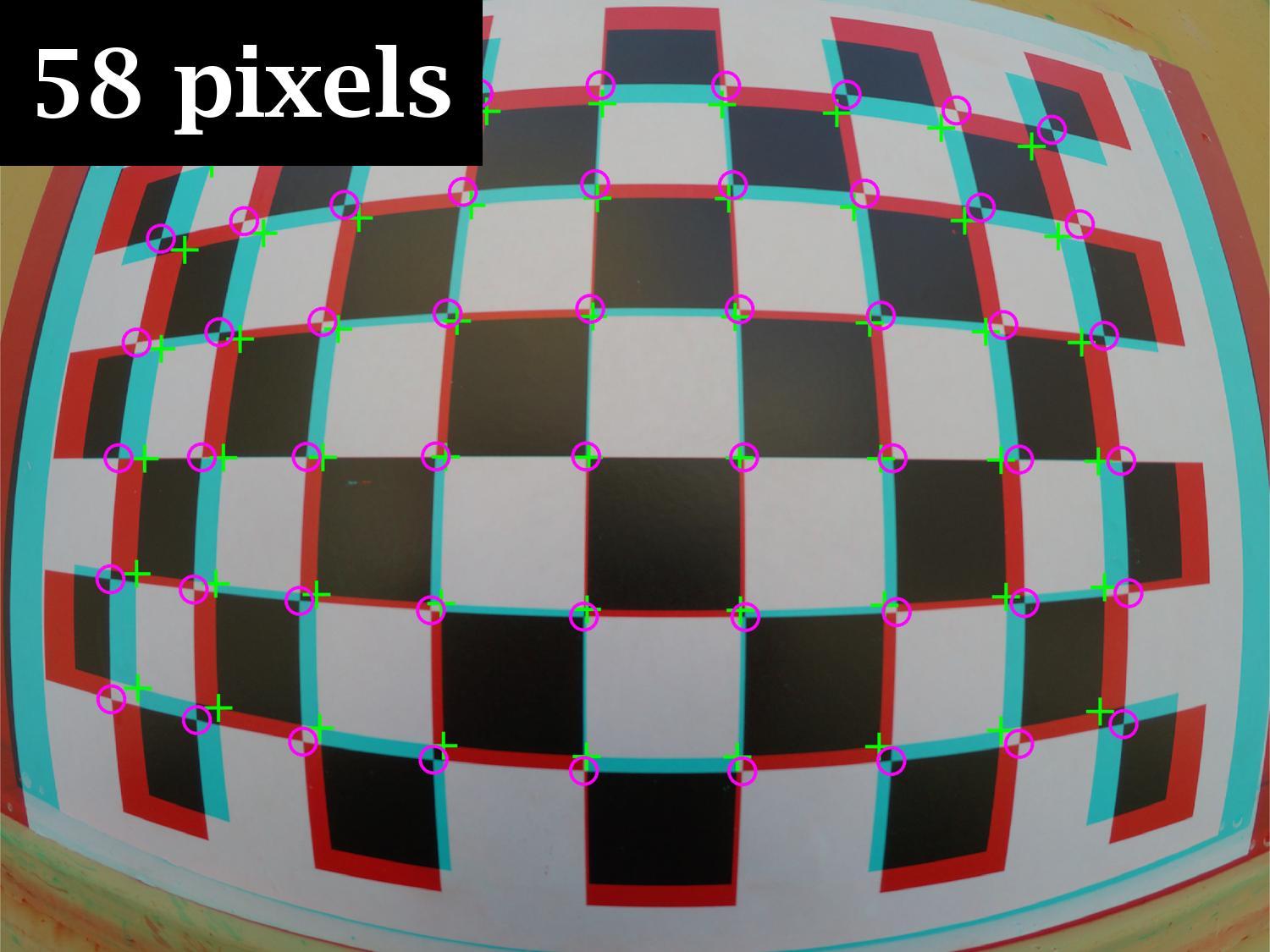} &
\includegraphics[height=2.34cm,valign=M]{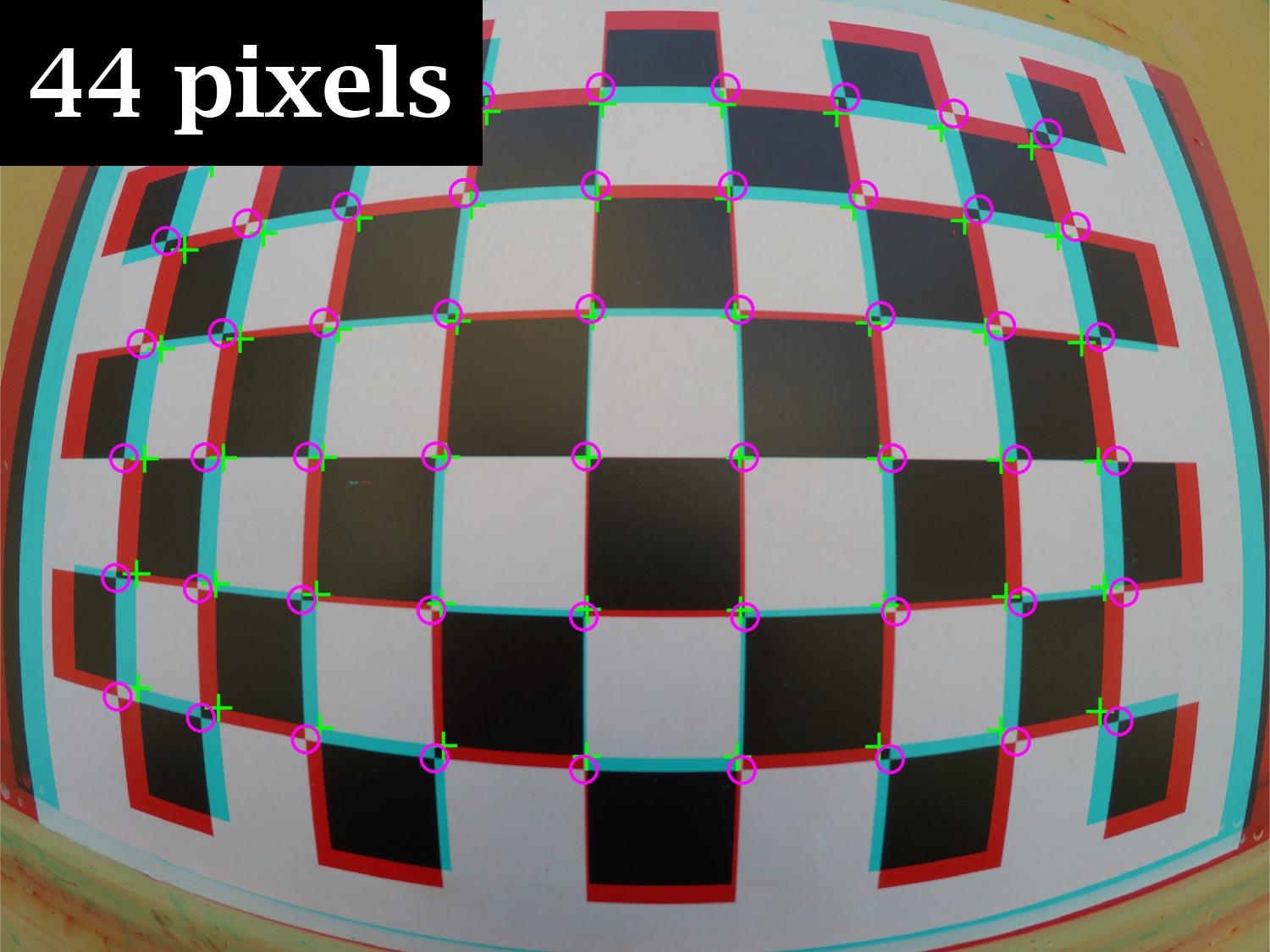} &
\includegraphics[height=2.34cm,valign=M]{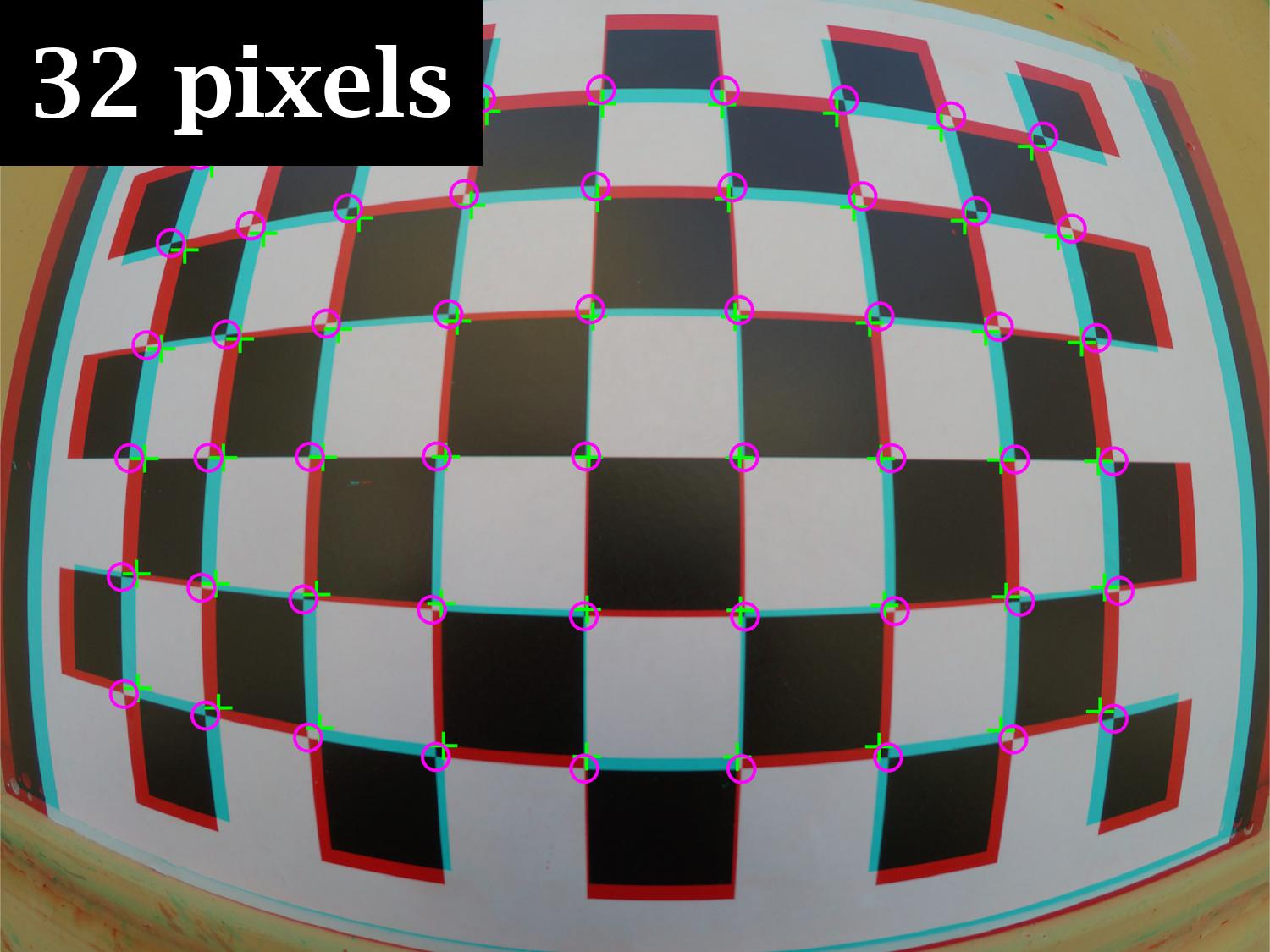} &
\includegraphics[height=2.34cm,valign=M]{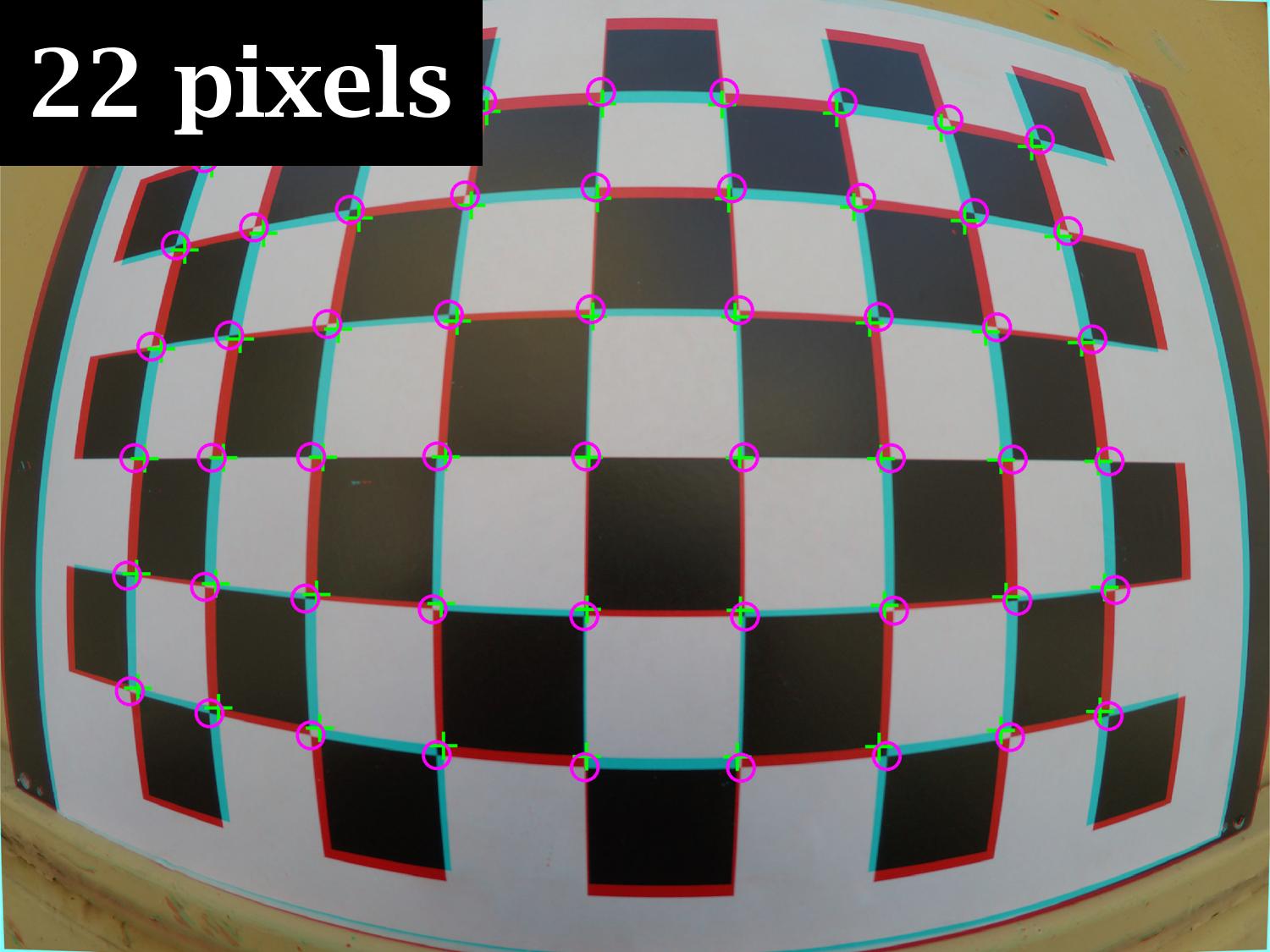} &
\includegraphics[height=2.34cm,valign=M]{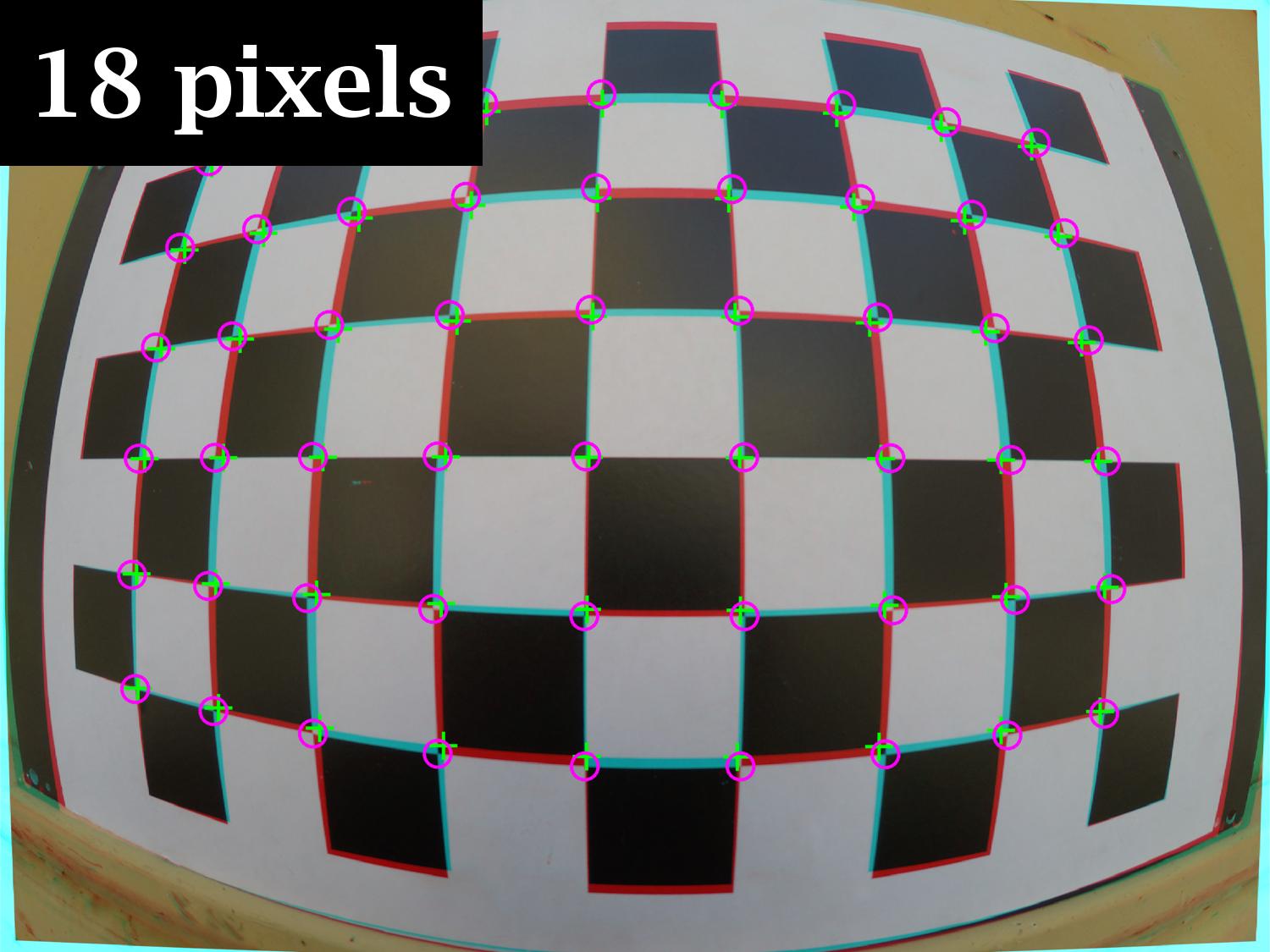} \\
& $10\%$ &
\includegraphics[height=2.34cm,valign=M]{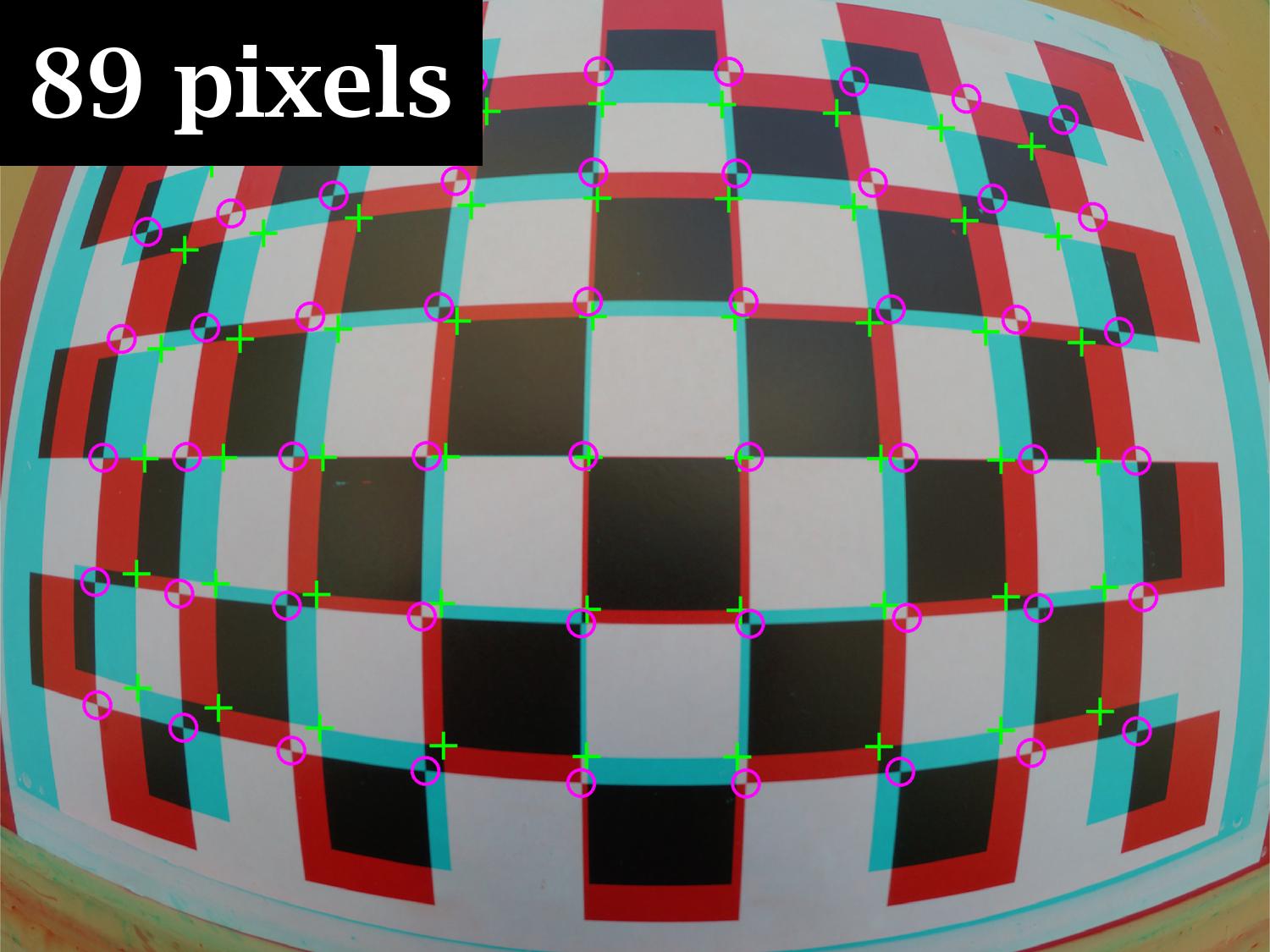} &
\includegraphics[height=2.34cm,valign=M]{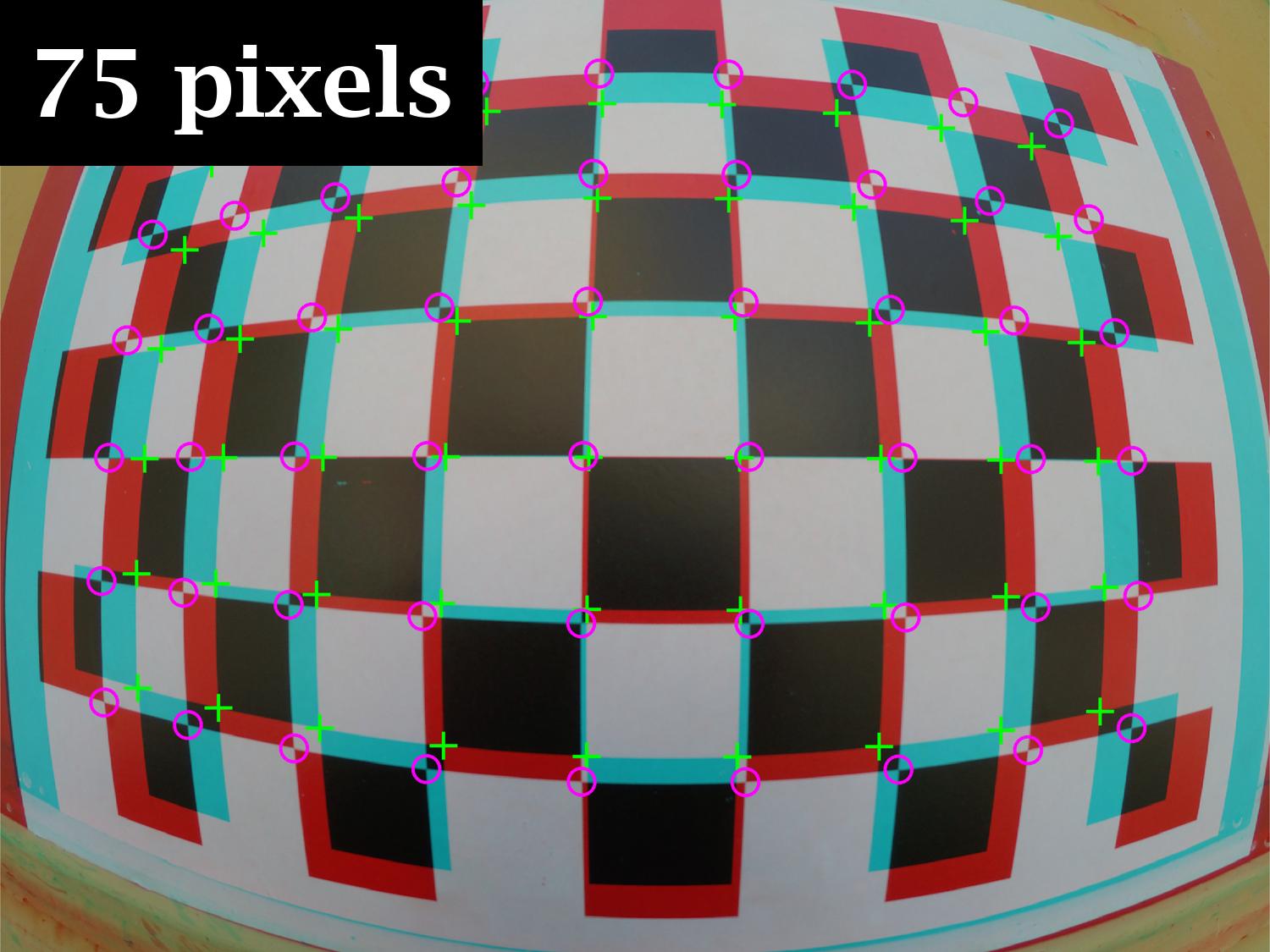} &
\includegraphics[height=2.34cm,valign=M]{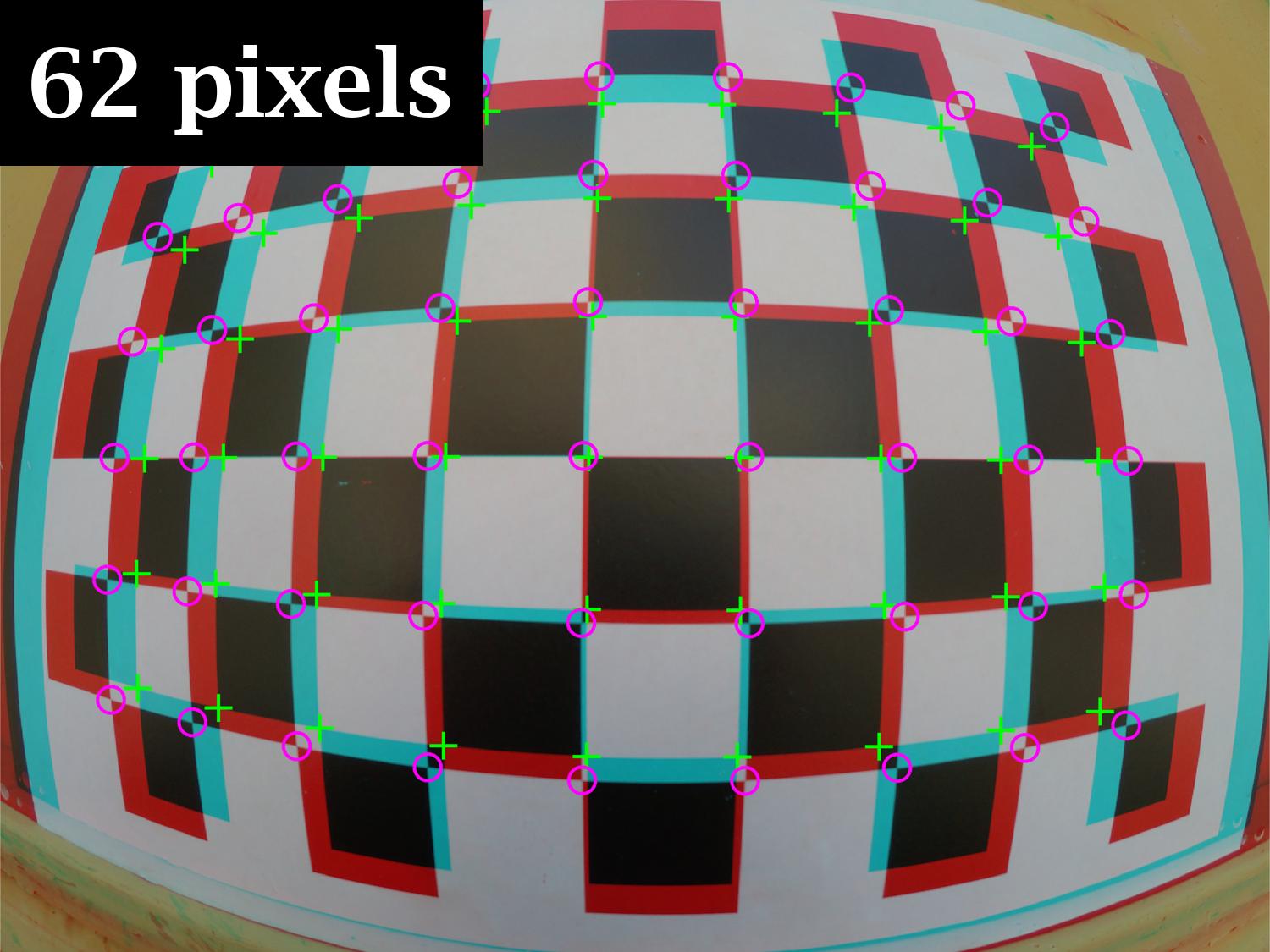} &
\includegraphics[height=2.34cm,valign=M]{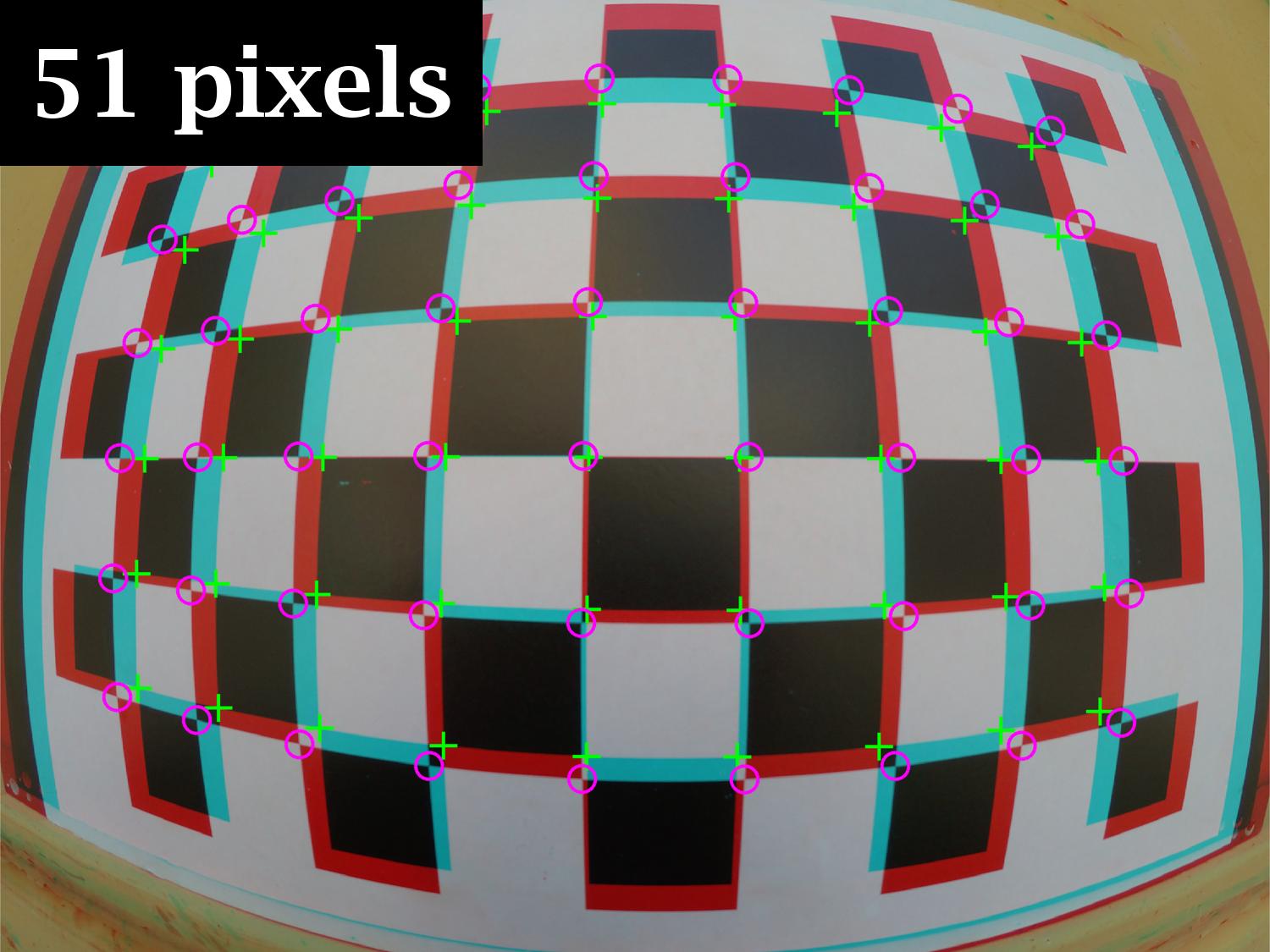} &
\includegraphics[height=2.34cm,valign=M]{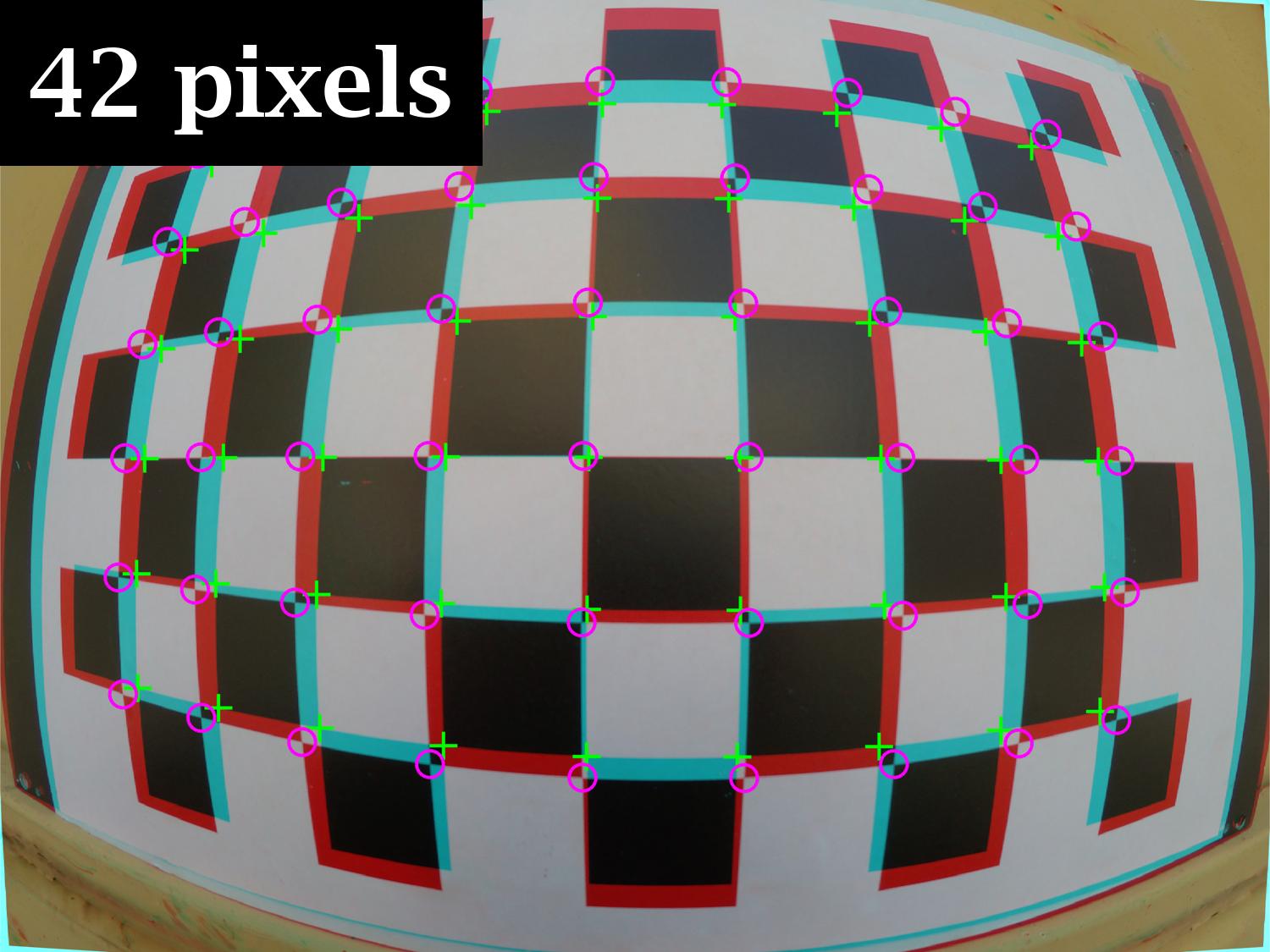}
\end{tabularx}
\mycaption{Intrinsics Error vs. Warp Error}{Each cell is the composite image of the original image with its red channel subtracted and the re-warped image with only its red channel. The RMS warp error computed for a combination of relative error levels of focal length and division-model parameter is rendered in the top left corner of each image. Errors from perturbed intrinsics are visible as false colors in the composite images, which highlight bad registrations between the original and re-warped images. The input image resolution is $3000\times2250$ pixels.} 
\label{tab:warperr}
\end{table*}

\newpage
\section{Relating Warp Error and Intrinsics}
\label{sec:supp_warp_error}
The warp error introduced in \secref{sec:warp_error} jointly
captures errors in the estimates of the focal length $f$ and the
division model parameter of lens undistortion $\lambda$. \figref{fig:warperr} and \tabref{tab:warperr} show the relation
between the errors in the estimates of intrinsics parameters and the
warp error. 
\figref{fig:warperr} shows that the warp error is nonlinear, nonsymmetric function of the intrinsic relative errors. An error in focal length can be compensated by an error in undistortion and vice-versa.

The chessboard images in \tabref{tab:warperr} give geometric intuition of how the metric warp error \rmswarperr corresponds to registration errors between the original image and the re-warped image as synthesized according to \secref{sec:warp_error}. The images confirm an observation that the warp error is not proportional in its arguments. \Eg $10\%$ relative error of undistortion with no error in focal length results in $25$ pixels \rmswarperr however increasing the relative error of focal length to $5\%$ will give $18$ pixels \rmswarperr which means that the algebraic errors partially compensate each other leading to the lower geometric error. 

\section{Additional Synthetic Experiments}
\label{sec:supp_AIT_stats}
\begin{figure*}[!t]
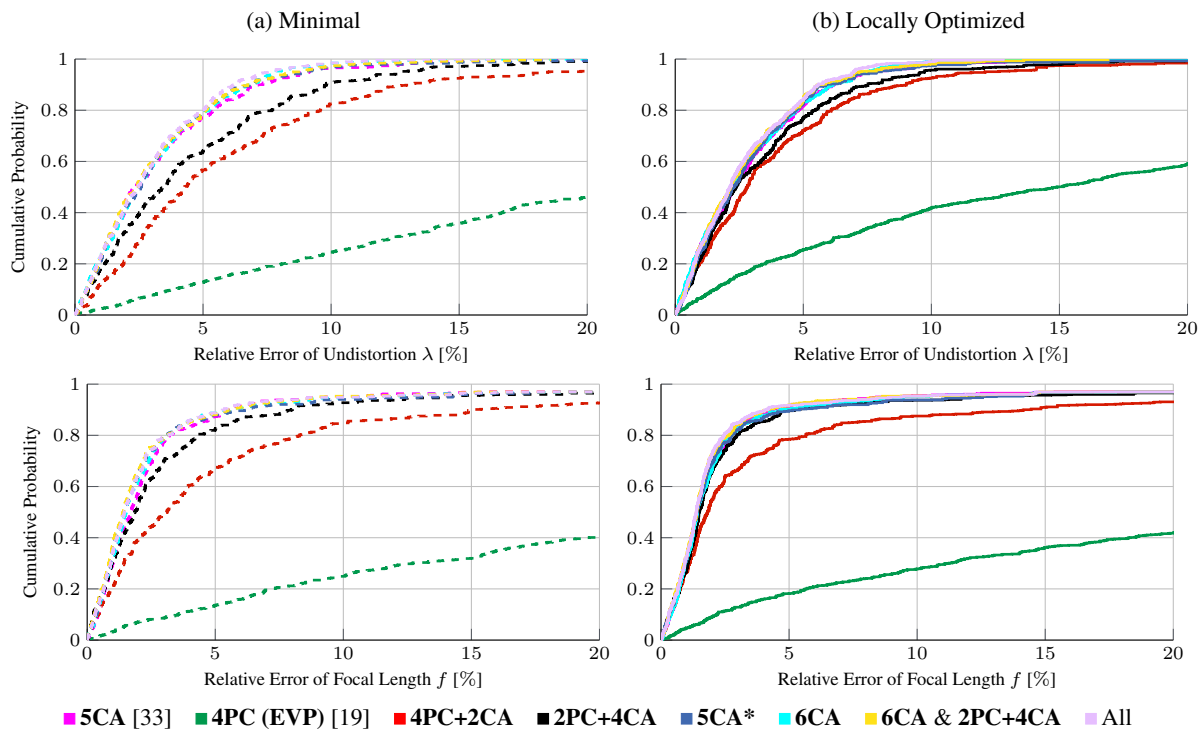

\centering
\setlength\fwidth{0.39\textwidth}
\subfloat[Minimal]{
\vspace{-7pt}
\scriptsize{\input{fig/AIT__min__relerr_q.tikz}}
\label{fig:supp_AIT_minimal}}\hspace{5pt}
\subfloat[Locally Optimized]{
\vspace{-7pt}
\scriptsize{\input{fig/AIT__relerr_q.tikz}}
\label{fig:supp_AIT_final}}\\
\scriptsize{\input{fig/AIT__min__relerr_f.tikz}}
\scriptsize{\input{fig/AIT__relerr_f.tikz}}\\
\small{\begin{tikzpicture}
\begin{customlegend}
[legend columns=-1,
legend style=
{draw=none,/tikz/every even column/.append style={column sep=0.25cm},cells={align=center}},
legend cell align={left},
legend entries={
\WildBMVC~\cite{Wildenauer-BMVC13},
\EVP~\cite{Pritts-CVPR18},
\HybridSolver,
\HybridSolverTwo,
\CircleDegSolver,
\CircleSolver,
\CircHyb,
\AllProposed
}
]
\addlegendimage{WildBMVC,fill=WildBMVC,only marks,mark=square*}
\addlegendimage{EVP,fill=EVP,only marks,mark=square*}
\addlegendimage{HybridSolver,fill=HybridSolver,only marks,mark=square*}
\addlegendimage{HybridSolverTwo,fill=HybridSolverTwo,only marks,mark=square*}
\addlegendimage{CircleDegSolver,fill=CircleDegSolver,only marks,mark=square*}
\addlegendimage{CircleSolver,fill=CircleSolver,only marks,mark=square*}
\addlegendimage{CircHyb,fill=CircHyb,only marks,mark=square*}
\addlegendimage{AllProposed,fill=AllProposed,only marks,mark=square*}
\end{customlegend}  
\end{tikzpicture}}
\caption{Cumulative distributions of the relative error of the division model parameter $\lambda$ and the relative error of the focal length $f$ on AIT dataset~\cite{Wildenauer-BMVC13}. Results are shown for (a) the minimal \ie initial solutions and (b) locally optimized \ie final solutions.}
\label{fig:supp_AIT_stats}
\vspace{-15pt}
\end{figure*}

\figref{fig:supp_AIT_stats} reports the cumulative distributions of the relative errors of undistortion and the relative errors of focal length for the individual solvers, the combination of all solvers, and the \CircHyb combination.

\vfill

\section{Additional Real-Image Experiments}
\label{sec:supp_real_data}
\figref{fig:supp_real_data} provides qualitative results for the proposed solvers. The images were taken with four lenses mounted on Canon 5DSR camera: Sigma 8mm, Samayang 12mm, Sigma 15mm and Sigma 24mm. The solvers accurately calibrate cameras with fields of view from narrow to fisheye with diverse image
content.

\newcommand{\addrows}[1]{
  \includegraphics[width=0.3\textwidth]{suppimg/#1/#1.jpg} &
  \includegraphics[width=0.3\textwidth]{suppimg/#1/#1__ud__CircleSolver+HybridSolverTwo.jpg} &
  \includegraphics[width=0.3\textwidth]{suppimg/#1/#1__rect1__CircleSolver+HybridSolverTwo.jpg}
}

\renewcommand{\tabcolsep}{3pt}
\newcolumntype{P}[1]{>{\centering\arraybackslash}p{#1}}
\begin{figure*}[!t]
\centering
\captionsetup[subfigure]{labelformat=empty}
{\def\arraystretch{0.7}
\begin{tabular}{P{1cm} P{5.2cm} P{5.2cm} P{5.2cm}}
Lens & Input & Undistorted & Rectified\\
  \rotatebox{90}{\hspace{22pt}Sigma 24mm}
& \addrows{8S8A0366}\\
  \rotatebox{90}{\hspace{18pt}Sigma 15mm}
& \addrows{8S8A0304}\\
  \rotatebox{90}{\hspace{12pt}Samayang 12mm}
& \addrows{8S8A0255}\\
  \rotatebox{90}{\hspace{22pt}Sigma 8mm}
& \addrows{8S8A0431}\\
\end{tabular}
}
\mycaption{Field of View Study}{Auto-calibration results are shown for lenses with different fields of view from narrow to fisheye. The minimal sample ---green circles and blue regions---of the returned solution is depicted on the input image.}
\vspace{-15pt}
\label{fig:supp_real_data}
\end{figure*}

\end{document}